\documentclass[conference]{IEEEtran}
\IEEEoverridecommandlockouts

\usepackage{amssymb, amsmath, mathrsfs}
\usepackage{graphicx,amsmath}
\usepackage{url,xspace}
\usepackage{textcomp}
\usepackage{color}
\usepackage{subfigure}
\usepackage{algorithm}
\usepackage{algorithmic}
\usepackage{verbatim}
\usepackage{relsize}
\usepackage{url}
\usepackage{multirow}
\usepackage{times}
\usepackage{wrapfig}
\usepackage{balance}

\def\baselinestretch{0.971}

\newtheorem{definition}{Definition}

\def\base#1{{\small\texttt{#1}}}

\def\M{\mathbf{M}}

\def\H{\mathcal{H}}
\def\L{\mathcal{L}}

\def\Q{Q}
\def\W{\mathcal{W}}

\def\<{\langle}
\def\>{\rangle}

\newcommand{\nop}[1]{}

\begin{document}

\title{Mining News Events from Comparable News Corpora: A Multi-Attribute Proximity Network Modeling Approach}

\author{\IEEEauthorblockN{$^1$Hyungsul Kim, $^1$Ahmed El-Kishky, $^2$Xiang Ren, $^1$Jiawei Han}
\IEEEauthorblockA{$^1$\textit{The University of Illinois at Urbana-Champaign}, $^2$\textit{University of Southern California} \\
 \{hkim21,elkishk2,hanj\}@illinois.edu, xiangren@usc.edu }
}

\IEEEoverridecommandlockouts
\IEEEpubid{\makebox[\columnwidth]{978-1-7281-0858-2/19/\$31.00~\copyright2019 IEEE \hfill} \hspace{\columnsep}\makebox[\columnwidth]{ }}
\maketitle
\IEEEpubidadjcol

\begin{abstract}
We present \textbf{ProxiModel}, a novel event mining framework for extracting high-quality structured event knowledge from large, redundant, and noisy news data sources. The proposed model differentiates itself from other approaches by modeling both the event correlation within each individual document as well as across the corpus. To facilitate this, we introduce the concept of a proximity-network, a novel space-efficient data structure to facilitate scalable event mining. This proximity network captures the corpus-level co-occurence statistics for candidate event descriptors, event attributes, as well as their connections. We probabilistically model the proximity network as a generative process with sparsity-inducing regularization. This allows us to efficiently and effectively extract high-quality and interpretable news events. Experiments on three different news corpora demonstrate that the proposed method is effective and robust at generating high-quality event descriptors and attributes. We briefly detail many interesting applications from our proposed framework such as news summarization, event tracking and multi-dimensional analysis on news. Finally, we explore a case study on visualizing the events for a Japan Tsunami news corpus and demonstrate ProxiModel's ability to automatically summarize emerging news events.
\end{abstract}

\section{Introduction}\label{sec:introduction}

With the proliferation of digital media and newswires, massive online news data has become widely available. Subsequently, automated analysis of news events has become an important research issue since the sheer quantity of news events makes human-powered analysis intractable~\cite{li2005probabilistic,zi2004news,tao2014newsnetexplorer}.

An interesting phenomenon within these large new corpora is that in addition to a large coverage of news events within a corpus, individual articles within a collection often contain redundant, partially overlapping content with each other. This overlapping content provides an opportunity to align articles and discover what is important.

More formally, this information redundancy from partially overlapping content across news articles provides the statistical power necessary to confidently identify and describe important events as well as their essential attributes such as time, location, and relevant persons and organizations. Moreover, because news articles often concurrently cover multiple related events, the vast redundancy facilitates the discovery of the connections that link events forming a comprehensive new event timeline.

Discovering, extracting, and visualizing events along with their key descriptors, attributes, and temporal and relevancy connections makes it possible to both understand the intricacies of an event while simultaneously forming a big-picture by linking events. As such, it's desirable to build a system that, given a news corpus, automatically discovers important events, attributes key properties to them accurately, and connects them thematically and temporally.

\begin{figure*}[t]
\centering
\includegraphics[width = 7.08in, angle=0]{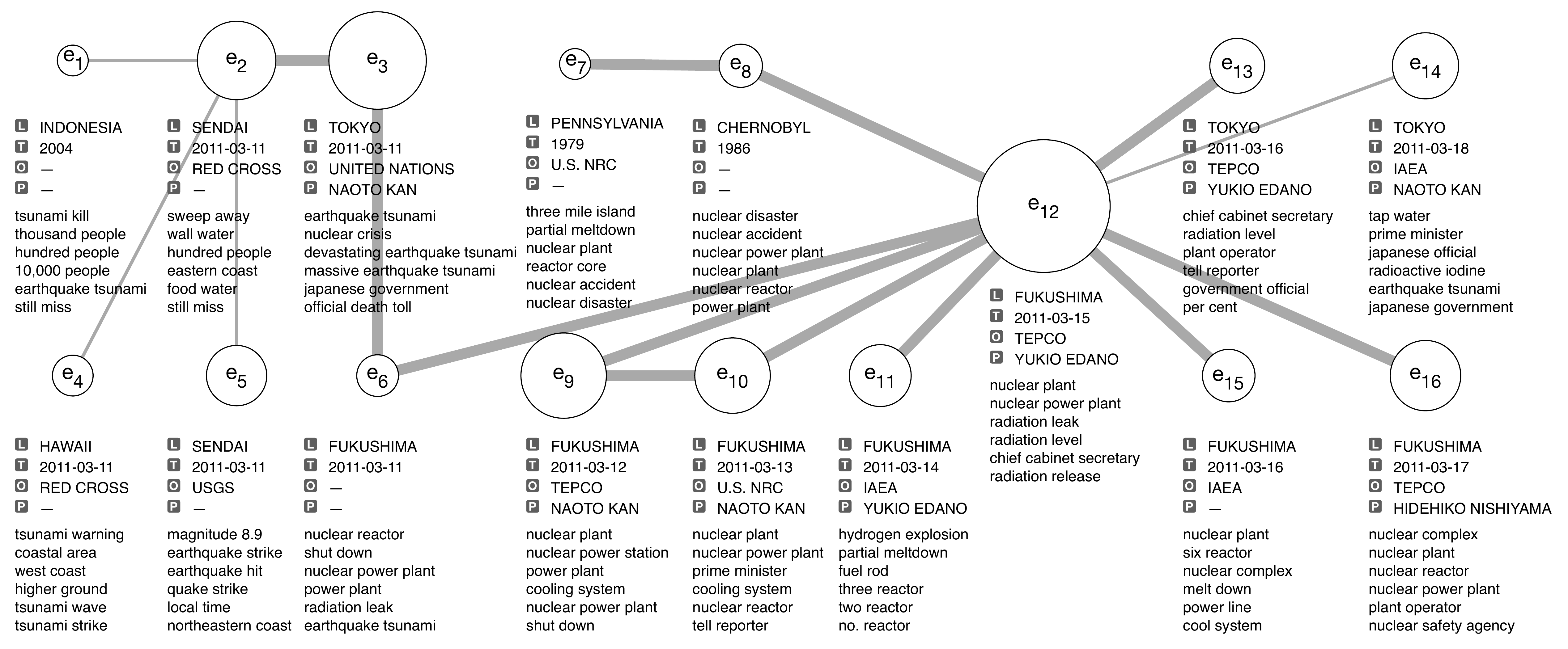}
\caption{Events, their attributes and event connections for the 2011 Japan tsunami and nuclear accident. Automatically generated by ProxiModel.
}
\label{fig:big-picture}
\vspace{-0.1in}
\end{figure*}


There have been multiple approaches to summarize and visualize news events. However, these approaches suffer from several limitations.
\begin{enumerate}
\parskip 0.4ex

\item \textbf{Unigram-based event descriptors}: While some systems use \emph{unigrams} (i.e., \emph{single words}) as event descriptors~\cite{li2005probabilistic, mei2006probabilistic}, it has been shown that \emph{phrases} are more descriptive and interpretable than words~\cite{wang2013phrase, elkishky2014topmine}. 
    There are several studies that use phrases in information flow detection on the Web~\cite{leskovec2009meme, suen2013nifty} or in event detection with micro blogs~\cite{ritter2012open,li2012tedas}.
    However, their phrases are not for describing events but for searching and linking multiple documents.  

\item \textbf{Lack of key attributes for event description}: A reader can better understand the context of an event if aware of several key attribute values of an event: \emph{when} and \emph{where} the event happened, and \emph{who} or \emph{which organizations} the event is related to ~\cite{wagner2009rich,ren2016automatic}. Most previous works do not provide such defining attributes to events. Others apply heuristics such as extracting event attributes from meta data like publication dates and reporting locations, which can be noisy and misleading. Other key values, such as persons or organizations, are often unavailable or inaccurate.

\item \textbf{Ignoring event connections within a single document}: Events naturally relate to each other. For example, in 2011, an earthquake off the coast of Japan triggered a tsunami; this tsunami propelled a series of incidents that led to the 2011 Japan nuclear disaster. While these connections are often explicitly addressed within news articles, many event detection and tracking studies in micro blogs~\cite{sakaki2010earthquake, li2012tedas, ritter2012open} and news articles~\cite{li2005probabilistic, mei2006probabilistic} make the strong assumption that each document describes a single event.
While this assumption may hold true for short documents such as micro blog posts, long documents like news articles are more susceptible to event drift and may contain multiple related events.
\end{enumerate}


Given the assumption that noisy news text corpora is plentiful and these corpora contain news articles of redundant and comparable content, we propose a model for (1) identifying key attributes and descriptors, (2) grouping them, and (3) denoising them to extract high-quality event knowledge. 
This new approach, ProxiModel (\textbf{Proxi}mity network-based generative \textbf{model}), leverages the notion of proximity: \emph{If two instances co-occur in news articles closely and frequently, they have high proximity}. ProxiModel leverages this notion of proximity to model events, descriptors, attributes, and connections.


Fig.~\ref{fig:big-picture} shows an example output of ProxiModel for a news collection covering the 2011 Japan tsunami and subsequent nuclear accident.
There are $16$ events shown, with automatically generated phrasal key descriptors and event attributes, where circle size represents the importance of events, and line width the strength of event connection.


By automatically identifying latent news events, their phrasal descriptors, attributes, and connections, ProxiModel provides an effective framework for organizing and exploring these huge amounts of data. This process is performed without understanding the semantic meaning of the content in the news articles. Instead, the method purely utilizes the interconnections between event attributes and their proximity.
ProxiModel possesses several key qualities that differentiate it from other event detection methods and allow for high-quality event discovery and intuitive and interpretable organization of news:
(1) it provides a big picture of events in news articles with rich information, which includes the importance of events, key phrasal descriptors, event attributes, and event connections, (2) it solely utilizes proximity information with regularized sparsity on model parameters to find correct event attributes and connections from text, and
(3) it uses a scalable data structure, called a proximity network, that stores necessary information from news articles.

The remaining portion of the paper is organized as follows:
Section~\ref{sec:background} introduces the preliminaries and definitions. Section~\ref{sec:method} describes our construction of proximity networks, followed by our generative models and the model learning process. Our experimental setup and results are described in Section~\ref{sec:experiments}.
The related work is discussed in Section~\ref{sec:related}, and Section~\ref{sec:conclusion} concludes our study.

\section{Preliminaries}\label{sec:background}

While bearing some similarities, event discovery has subtle differences from topic discovery or topic modeling. Traditionally, a topic is defined as a distribution of words~\cite{blei2003latent}. An event, however, is associated with several key attributes including location, time, person, organization, and a set of descriptive phrases to denote theme.

We first examine several key attributes and primitives of events.

\begin{figure}[t]
\centering
\includegraphics[width = 3.2in, angle=0]{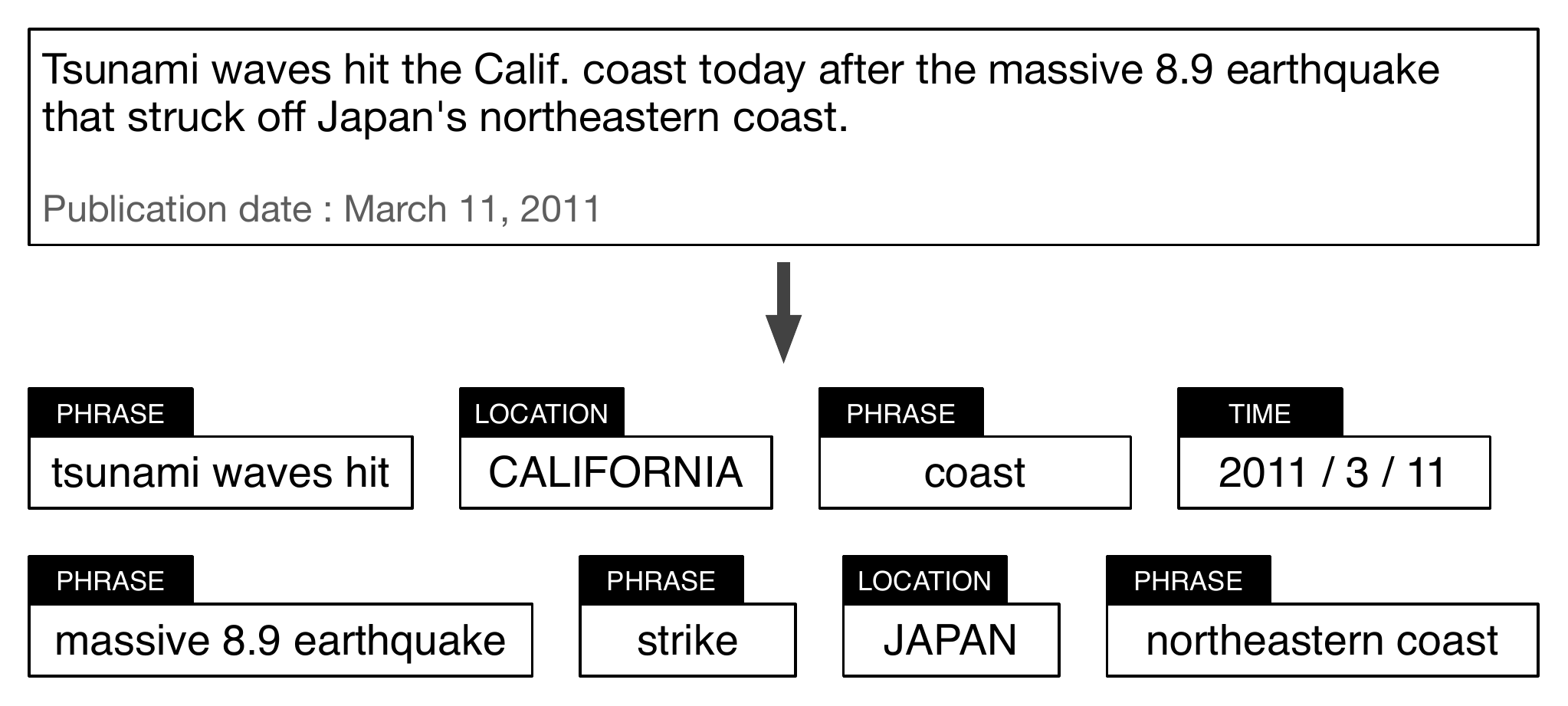}
\vspace{-0.1in}
\caption{The representation of a document as a sequence of bases. NLP tools and a phrase mining algorithm are used to segment documents.}
\label{fig:segmentation}
\vspace{-0.1in}
\end{figure}

\smallskip
\noindent
\textbf{1. Time}: Temporal expressions are extracted from documents and normalized to the form of the TIMEX3, which is a part of the TimeML annotation language~\cite{pustejovsky2003timeml}. Relative temporal expressions like ``last night'' and ``yesterday'' are also normalized by taking the report time or publication time of the document as the fixed reference time. For example, the word ``today'' in Figure~\ref{fig:segmentation} is mapped to ``2011/3/11'' because of the publication date. In this paper, we informally refer to the extracted normalized time expressions as time.

\smallskip
\noindent
\textbf{2. Location}:
Locations are geo-political entities such as city, state, and country. They are extracted and normalized to their surface forms. For example, the word ``Calif.'' is mapped to ``CALIFORNIA'' in Figure~\ref{fig:segmentation}.

\smallskip
\noindent
\textbf{3. Person}
Extracted persons are not only public figures, but also private figures who are mentioned in news articles. For example, Jun-seok Lee, who was the captain of the sunken Sewol Ferry, is extracted. Co-references are also resolved within a document (e.g., Captain Lee is mapped to Jun-seok Lee).

\smallskip
\noindent
\textbf{4. Organization}
Companies, governments, and other organizations are extracted. An abbreviation of an organization is mapped to its full name if they are mentioned in the same document. For example, TEPCO is mapped to Tokyo Electric Power Company.

\smallskip
\noindent
\textbf{5. (Thematic) phrases}:  A phrase is a sequence of contiguous words that represents a meaningful semantic unit.
Recently developed phrase mining algorithms such as ToPMine~\cite{elkishky2014topmine} and SegPhrase~\cite{liu2015segphrase} perform fast, pruning-based frequent contiguous pattern mining and then statistically reason about the significance of the contiguous co-occurrence while applying context constraints to discover meaningful phrases.  
We use ToPMine~\cite{elkishky2014topmine} to mine quality phrases representing the above dimensions as well as thematic phrases that form a thematic dimension as shown in
Fig.~\ref{fig:segmentation}.

\smallskip
For simplicity, we refer to any phrase, time, location, person, or organization as a \emph{basis}.
A document $d$ is a sequence of segments $\langle d_1, d_2, \dots, d_k \rangle$, where $d_i$ corresponds to a basis. The order of segments corresponds to the order of original word tokens in the document. For a given comparable news corpus, we want to discover events defined as follows:

\begin{definition}[Event]
An event, $z$, is a real-world occurrence represented as a 5-tuple $\langle \eta_z, \phi^L_z, \phi^T_z, \phi^O_z, \phi^P_z \rangle$, where
$\eta_z$ is the distribution over all phrases,
$\phi^L_z$ is that (distribution) over all locations,
$\phi^T_z$ is that over all time,
$\phi^O_z$ is that over all organizations,
and $\phi^P_z$ is that over all persons.
\end{definition}




\section{Event Mining}\label{sec:method}

In this section, we introduce the concept of a proximity network and demonstrate the process of generating one from a comparable news corpus. We then propose an event mining method that operates efficiently on the constructed proximity network.


\subsection{Comparable News Corpus}

\begin{figure}[t]
\centering
\includegraphics[width = 3.4in, angle=0]{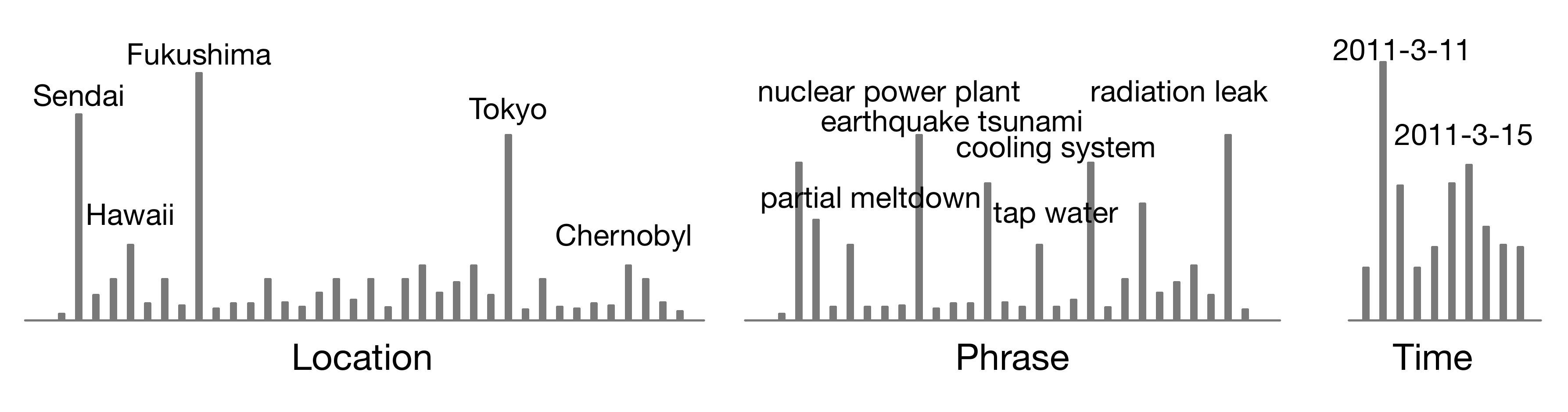}
\vspace{-0.25in}
\caption{Statistical power of comparable news corpus: key information can be easily discovered by counting the occurrences of basis}
\label{fig:comparable-corpora-power}
\vspace{-0.05in}
\end{figure}

In NLP tasks such as machine translation, a comparable corpus is a corpus consisting of documents on similar topics~~\cite{baker1995corpora}. Similarly, we define a comparable news corpus is a collection of news articles that cover related events. Manually obtaining such corpora is relatively simple and can be done, for example, by simply using keyword search on a news database. Given that such a corpus is curated using a few key-word searches, the resultant comparable news corpus contains many documents with partially repeated information and common phrases for important events. These documents with their overlapping pieces of information can aid in analyzing and understanding the underlying events that these documents detail.

Here we briefly illustrate the potential of a collective analysis on a comparable news corpus, with two simple but incomplete analysis methods.
The first is counting the occurrences of key attributes such as locations, phrases, and time as shown in Fig.~\ref{fig:comparable-corpora-power}. The second is by counting redundant information across the news articles about Japan tsunami in 2011, the peaks show important information in each dimension.

Unfortunately, such peaks, generated from document-level co-occurrences of key attributes may be inaccurate and are as such unsuitable for extracting events. For example, a hydrogen explosion in a nuclear power plant happened in Fukushima on March 14, 2011. The phrase ``hydrogen explosion'', however, has high co-occurrence with ``2011-03-11'' because most of the news articles mentioned the earthquake on March 11, 2011 to address the cause of the damaged nuclear power plant. To avoid these problems, events should be resolved by considering the \emph{proximity} of bases within documents.

\subsection{Proximity Network}\label{sec:network}

\begin{figure}
\centering
\includegraphics[width=3.5in, angle=0]{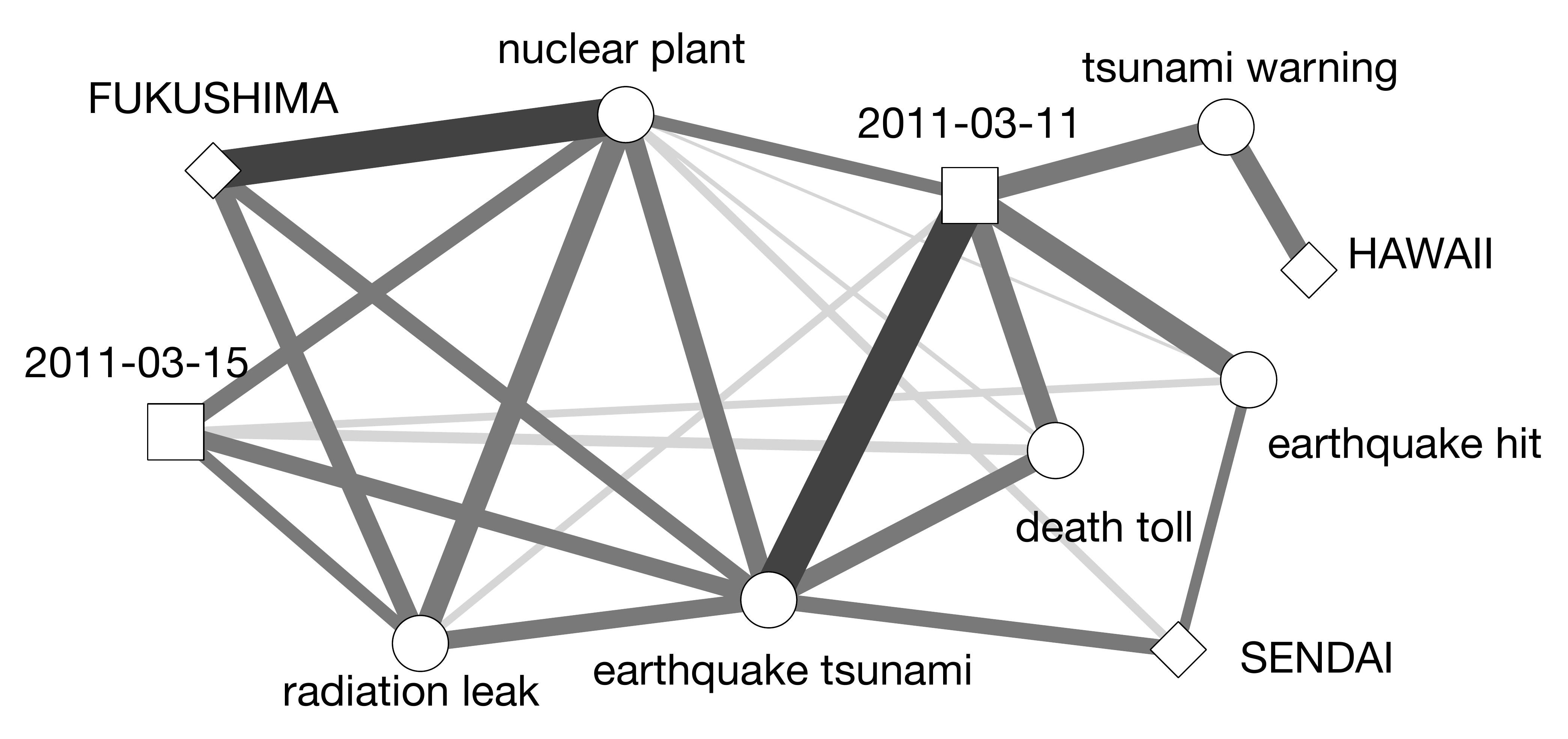}
\vspace{-0.25in}
\caption{An example of proximity networks from the Japan Tsunami corpus.  There are three types of nodes: $\bigcirc$: a phrase node, $\Diamond$: a location node, and $\Box$: a time node. Line thickness indicates the weight of the corresponding edge in log scale.}
\label{fig:toy-network}
\vspace{-0.05in}
\end{figure}

Proximity is a measure of how close two terms occur in a document or a passage. This measure has been successfully adopted in many different tasks including word association~\cite{washtell2009co, washtell2009comparison}, document retrieval~\cite{lv2009positional, zhao2011crter}, named entity retrieval and expert finding~\cite{balog2009language, petkova2007proximity}.

Proximity is an important cue for estimating the strength of association between two bases, in which a strong association between two bases indicates they belong to the same event. For example, in the Japan Tsunami news corpus, we find time expressions of \base{2011/03/11} near \base{earthquake hit} phrase frequently: This is the time when a massive earthquake hit Japan. In addition, we find the phrase \base{radiation leak} around mentions of the location Fukushima more often than other locations: Similarly, Fukushima is the city where crippled nuclear power plants had radiation leaks.

We want to collect such evidence or associations between bases in an efficient way by constructing an information network that we refer to as a \textit{proximity network}. We define a proximity network as a heterogenous network with different types of nodes and edges between them. The set of nodes in the proximity network is the set of bases in a given corpus $\mathcal{C}$, and the weight of an edge between two nodes is based on proximity between the nodes as follows.

$$ e_{x, y} = \sum_{ d \in \mathcal{C} } \sum_{1 \leq i < j \leq N_d} \delta_d(i, x) \delta_d(j, y) k(i, j),$$
where $\delta_d : \mathbb{N} \times \mathcal{B} \rightarrow \{0, 1\}$ is an indicator function and $k(i, j)$ is a proximity kernel such that
\begin{align*}
\delta_d(i, x) &=
\left\{
	\begin{array}{ll}
		1 & \mbox{if segment at position $i$ in $d$ corresponds to $x$}\\
		0 & \mbox{otherwise}
	\end{array}
\right. \\
k(i, j) &= exp \left[ \frac{-(i-j)^2}{2\sigma^2} \right]
\end{align*}


Note that $\sigma$ is a constant that controls the propagation scope of each segment. A proximity network with small $\sigma$ captures very different information from one with large $\sigma$. We will use two proximity networks with different $\sigma$s to model different information as discussed in the next section.

An example of proximity networks is shown in Figure~\ref{fig:toy-network}, generated from the Japan Tsunami corpus with $\sigma = 1$. It shows strong proximity between \base{FUKUSHIMA} and \base{nuclear plant} and between \base{earthquake tsunami} and \base{2011-03-11}. If one tries to cluster the nodes in the figure based on the edges, such clustering may yield three clusters as follows:
\begin{enumerate}
\parskip -0.2ex
	\item \{ \base{2011-03-15}, \base{FUKUSHIMA}, \base{radiation leak}, \base{earthquake tsunami}, \base{nuclear plant} \}
	\item \{ \base{2011-03-11}, \base{SENDAI}, \base{earthquake tsunami}, \base{death toll}, \base{earthquake hit} \}
	\item \{ \base{2011-03-11}, \base{HAWAII}, \base{tsunami warning} \}
\end{enumerate}

We posit that some latent parameters guide the formation of clusters of nodes. As such, we model such latent parameters as events and describe how to best infer these latent events in the following section.

A proximity network constructed from the corpus could be noisy and dense without post-processing. Since our corpus has partially repeated news articles and important links get greater weights, we use link minimum support ($l_{minsup}$) to remove infrequent links (i.e., whose weights are less than $l_{minsup}$). This truncation not only removes noise in the network, but also makes the network sparse, where modeling becomes more efficient in relation to time and space.

\subsection{Proximity Network Generative Models}\label{subsec:model}

In this section, we describe the \emph{Proximity Network Generative Model} (ProxiModel). Proximity networks have pairwise proximity information among bases. Unlike previous studies that use heuristic proximity metrics~\cite{washtell2009co, washtell2009comparison, lv2009positional, zhao2011crter, petkova2007proximity}, we learn latent parameters from proximity to model events. Specifically, we design a probabilistic model for proximity networks to model events, in which edges in the networks are created according to a generative process. In order to model events with descriptors, attributes, and connections, we construct two proximity networks $N_s$ and $N_l$, with small $\sigma_s$ and large $\sigma_l$ values, from an input corpus.

\smallskip
\noindent
\textbf{Proximity Network $\boldsymbol{N_s}$}: $\sigma_s$ is set smaller than $\sigma_l$ to capture proximity within a smaller propagation scope. This proximity network is mainly used to learn event descriptors and attributes. It only has edges with at least one phrase end node. In other words, it only has edges consisting of
\base{phrase}-\base{phrase},
\base{phrase}-\base{time},
\base{phrase}-\base{location},
\base{phrase}-\base{organization}, or
\base{phrase}-\base{person}.

\smallskip
\noindent
\textbf{Proximity Network $\boldsymbol{N_l}$}: $\sigma_l$ is set greater than $\sigma_s$ to capture proximity within a larger propagation scope.
This proximity network is mainly used to learn event connections. It only has edges with two phrase end nodes. In other words, it has only edges of \base{phrase}-\base{phrase}.

\textbf{Our Assumptions}: In the generative model, we encode our assumptions as follows:
\begin{enumerate}
\parskip -0.2ex
\item Two phrases for the same event have high proximity in $N_s$. \label{st:two_phrases_sigma_1}
\item A phrase and an event attribute for the same event have high proximity in $N_s$. \label{st:one_phrase_sigma_1}
\item Two phrases from different events have high proximity in $N_l$ if the events are connected \label{st:two_phrases_sigma_2}
\item Each event has a few event attributes of the same type. \label{st:sparse_attribute}
\item There are only a few event connections. \label{st:sparse_connection}
\end{enumerate}
Note that two phrases for the same event have high proximity in $N_l$ as well as in $N_s$ because of the Gaussian kernel.

\begin{figure}[t]
\centering
\includegraphics[width = 3.4in, angle=0]{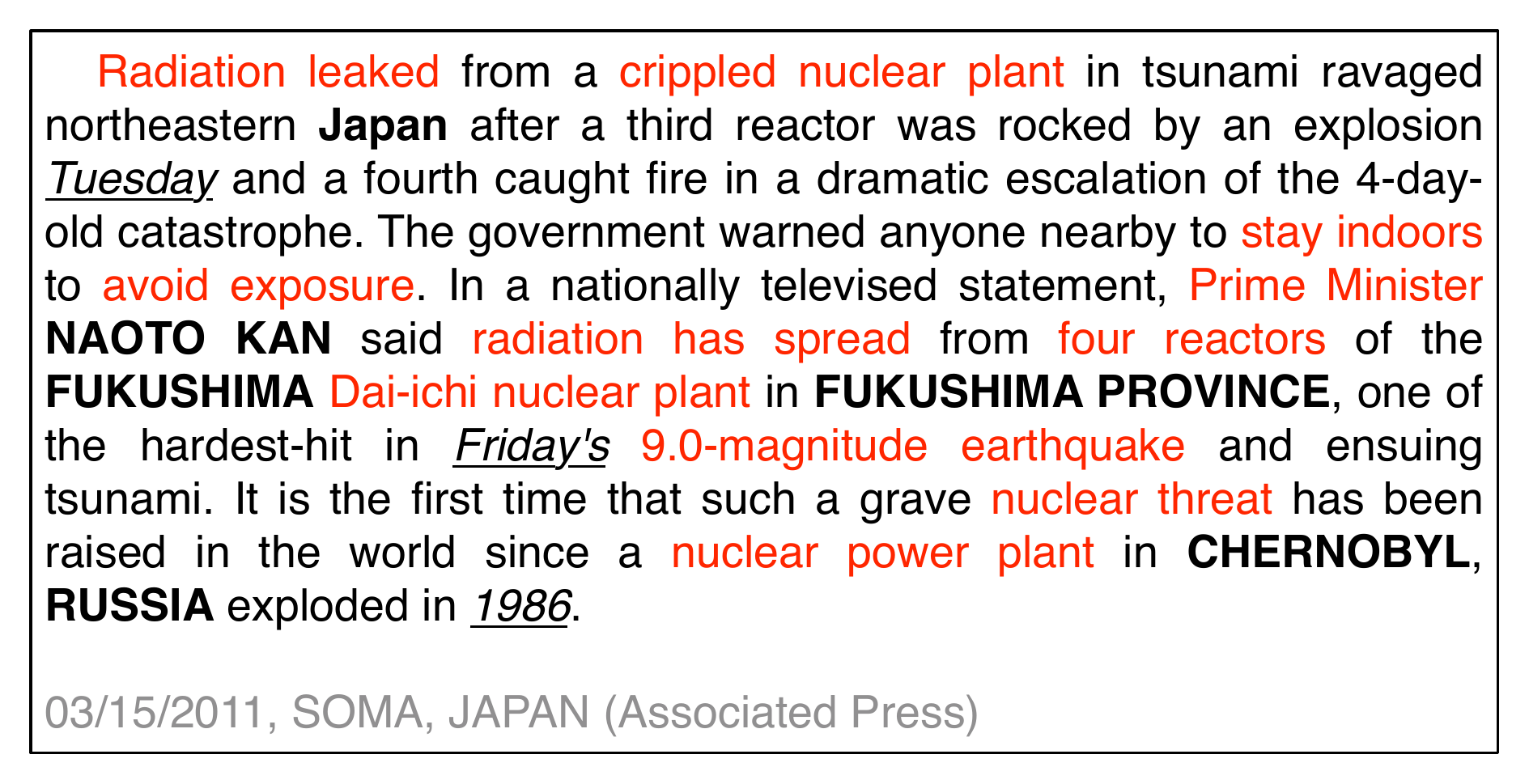}
\vspace{-0.15in}
\caption{An example news article: phrases are in red, named entities are in bold, and temporal expressions are in italic and underlined.}
\label{fig:news-example}
\vspace{-0.05in}
\end{figure}

We first address the assumptions with an example news article in Figure~\ref{fig:news-example}.
The news article mainly reports the leaked radiation from a crippled nuclear power plant in Fukushima, Japan, which happened on March 15, 2011.

The article also mentions a main cause of the damages in the nuclear power plant---a massive earthquake hit Japan in March 11, 2011 which caused strong tsunamis that damaged the nuclear power plant.
For example, \base{radiation leaked} and \base{crippled nuclear plant} have high proximity in $N_s$ as an example of Assumption~\ref{st:two_phrases_sigma_1}.
\base{9.0-magnitude earthquake} and \base{Friday} have high proximity in $N_s$ as an example of Assumption~\ref{st:one_phrase_sigma_1}.
In addition, \base{9.0-magnitude earthquake} and \base{crippled nuclear plant} have high proximity in $N_l$.

\subsubsection{Generative Process}
In our generative model, we convert the edge weights in $N_s$ and $N_l$ to multigraphs as follows: 
The number of edges between two nodes is equal to the integer part of the weight in the original network.
We denote the total number of edges in $N_s$ and $N_l$ by $n_s$ and $n_l$ respectively.

\begin{figure}[t]
\centering
\includegraphics[width=3.5in, angle=0]{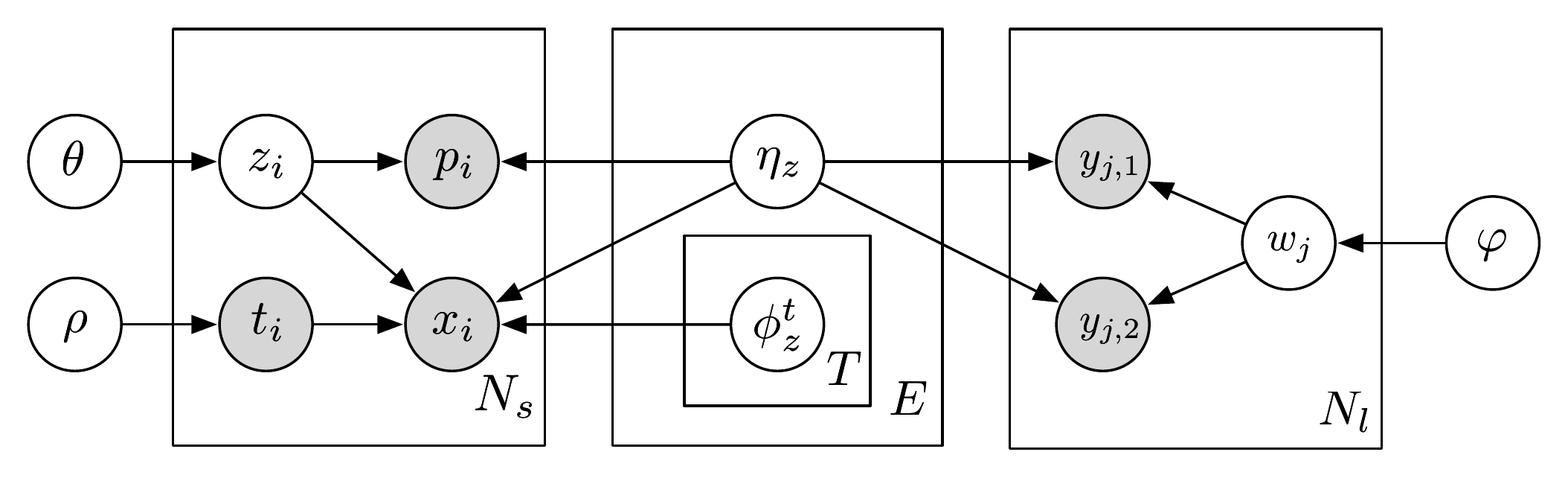}
\vspace{-0.25in}
\caption{A generative model for $\sigma_s$-proximity network($N_s$) and $\sigma_l$-proximity network($N_l$) }
\label{fig:graphical_representation}
\vspace{-0.05in}
\end{figure}

We define a generative process for edges in $N_s$ and $N_l$ as shown in Algorithm~\ref{algo:gen_process}.
In $N_s$, each edge belongs to one event, indicating two end points belong to the event. In $N_l$, end points of each edge can belong to different events as well as the same event.
See Figure~\ref{fig:graphical_representation} for a graphical representation of the model.

In this generative model, we can derive the distribution of the number of edges between any two nodes in $N_s$.
Generating an edge between a phrase-$i$ node and an attribute-$j$ node of type $t$ in event $z$ can be modeled as a Bernoulli trial with a success probability of $\theta_z \rho^t \eta_{z,i} \phi^t_{z,j}$. When $n_s$ is large, the total number of successes, $e^{s,t}_{i, j, z}$ asymptotically follows a Poisson distribution~\cite{serfling1978some} as follows:
$$
e^{s,t}_{i, j, z} \sim Poisson(n_s \theta_z \rho^t \eta_{z,i} \phi^t_{z,j}).
$$

Due to the additive property of Poisson distribution, we can derive that the observed variable $e^{s,t}_{i,j}$ follows a Poisson distribution as follows:
$$
e^{s,t}_{i,j} = \sum_{z} e^{s,t}_{i,j,z} \sim Poisson(\sum_{z} n_s \theta_z \rho^t \eta_{z,i} \phi^t_{z,j}).
$$

Thus, given the model parameters, the probability of all observed edges in $N_s$ is

$$
\L_s = p(\{e^{s,t}_{i,j}\} | \theta, \rho, \eta, \phi) = \prod_{i,j,t} \frac { (\mu^{i,j,t}_s) ^{e^{s,t}_{i,j}} exp(-\mu_s^{i,j,t}) } { e^{s,t}_{i,j}! },
$$
where $\mu_s^{i,j,t} = \sum_{z} n_s \theta_z \rho^t \eta_{z,i} \phi^t_{z,j}$.

Similarly, we can derive the distribution of the number of edges between any two nodes in $N_l$.

$$
e^{l}_{i,j} = \sum_{z_1, z_2} e^l_{i, j, z_1, z_2} \sim Poisson(\sum_{z_1,z_2} n_l \varphi_{z_1,z_2} \eta_{z_1,i} \eta_{z_2, j}).
$$

Thus, given the model parameters, the probability of all observed edges in $N_l$ is
$$
\L_l = p(\{e^l_{i,j}\} | \varphi, \eta) = \prod_{i,j} \frac { (\mu_l^{i,j})^{e^l_{i,j}} exp(-\mu_l^{i,j}) } { e^l_{i,j}! },
$$
where $\mu_l^{i,j} = \sum_{z_1,z_2} n_l \varphi_{z_1,z_2} \eta_{z_1,i} \eta_{z_2,j}$.

The overall probability of all observed edges in $N_s$ and $N_l$ is
$$
\L = \L_s \cdot \L_l.
$$

We encode Assumptions~\ref{st:two_phrases_sigma_1} and \ref{st:one_phrase_sigma_1} in the generative process of $N_s$, and Assumption~\ref{st:two_phrases_sigma_2} in the generative process of $N_l$.

To model the assumptions that each event has only a few event attributes and there are only few event connections, we introduce sparse regularization on model parameters as priors.

We impose an \emph{apriori} probability on the parameters given by
\begin{align}
\L' \propto \L \cdot p(\phi) \cdot p(\varphi),
\label{eq:likelihood}
\end{align}
where $p(\phi) = e^{-\sum_z \sum_t \alpha_t \H(\phi^t_z)}$, $p(\varphi) = e^{-\beta \H(\varphi)}$, $\H(x)$ is the Shannon's entropy of distribution $x$, and $\alpha_t$ and $\beta$ are sparse prior weights.
With higher values of $\alpha_t$ and $\beta$, event attributes and connections have lower entropy, i.e. are sparser.


\subsubsection{Parameter Learning}\label{subsec:model_learning}

We learn the model parameters by the Maximum Likelihood (ML) principle.
To deal with the normalization constants of the prior probabilities, the log-likelihood of Eq~(\ref{eq:likelihood}) must be augmented by appropriate Lagrange multipliers:
$ \Q = \log \L' + \textstyle \lambda_\theta \left(\sum_z \theta_z - 1 \right) + \lambda_\rho \left( \sum_t \rho_t - 1 \right) + \sum_{z} \lambda^z_\eta \left(\sum_i \eta_{z, i} - 1\right)
+ \textstyle \sum_{t,z} \lambda^{t,z}_\phi \left(\sum_i \phi^t_{z,i} - 1 \right) + \lambda_\varphi \left( \sum_{z_1,z_2} \varphi_{z_1,z_2} - 1 \right)
$

Then, we maximize $\Q$ using an Expectation-Maximization (EM) algorithm that iteratively infers the model parameters.

The E-step calculates the expected number of edges:
\begin{align}
\hat{e}^{s,t}_{i,j,z} &= e^{s,t}_{i,j} \frac{\theta_z \eta_{z,i} \phi^t_{z,j}}{\sum_k \theta_k \eta_{k,i} \phi^t_{k,j} } \\
\hat{e}^{l}_{i,j,z_1,z_2} &= e^{l}_{i,j} \frac{\varphi_{z_1,z_2} \eta_{z_1} \eta_{z_2}}{\sum_{k_1, k_2} \varphi_{k_1, k_2} \eta_{k_1} \eta_{k_2}}
\end{align}

In the M-step, the update equations for $\theta_z$, $\rho_t$, and $\eta_{z,i}$ are given by
\begin{align}
\theta_z = \frac{ \sum_{i, j, t} \hat{e}^{s,t}_{i,j,z} }{n_s}, \hspace{0.1in}
\rho_t = \frac{\sum_{i, j, z} \hat{e}^{s,t}_{i,j,z} }{n_s},\\
\eta_{z,i} = \frac{\sum_{j, t} \hat{e}^{s,t}_{i,j,z} + \sum_{j, z_2} \hat{e}^{l}_{i, j, z, z_2}}{\sum_{k, j, t} \hat{e}^{s,t}_{k,j,z} + \sum_{k, j, z_2} \hat{e}^{l}_{k, j, z, z_2}}
\end{align}

In the M-step, maximization of $Q$ with respect to $\phi$ and $\varphi$ leads to different sets of equations due to their priors and Lagrange multipliers:
\begin{align}
\frac{1}{\phi^t_{z,i}}\sum_j \hat{e}^{s,t}_{i,j,z} - n_s \theta_z + \alpha_t \log \phi^t_{z,i} + \alpha_t + \lambda^{t,z}_\phi = 0 \label{eq:phi}\\
\frac{1}{\varphi_{z_1,z_2}}\sum_{i,j} \hat{e}^l_{i,j,z_1,z_2} - n_l + \beta \log \varphi_{z_1,z_2} + \beta + \lambda_\varphi = 0.
\end{align}

The above set of simultaneous transcendental equations for $\phi$ and $\varphi$ can be solved using the Lambert's $\W$ function similar to~\cite{shashanka2008sparse}.
\begin{align}
\phi^t_{z,i} = \frac{-\sum_{j} \hat{e}^{s,t}_{i,j,z}/\alpha_t}{\W (-\sum_{j} \hat{e}^{s,t}_{i,j,z} e^{1 - n_s \theta_z / \alpha_t + \lambda^{t,z}_\phi / \alpha_t} / \alpha_t)}, \label{eq:lambert_phi}
\end{align}
where equations Eq.~(\ref{eq:phi}) and Eq.~(\ref{eq:lambert_phi}) form a set of fixed-point iterations for $\lambda^{t,z}_\phi$, and thus the M-step for finding $\phi^t_{z,i}$.

Similarly, we can get the following update equation for $\varphi_{z_1, z_2}$:
\begin{align}
\varphi_{z_1,z_2} = \frac{-\sum_{i,j} \hat{e}^l_{i,j,z_1,z_2}/\beta}{\W (-\sum_{i,j} \hat{e}^l_{i,j,z_1,z_2} e^{1 - n_l / \beta + \lambda_\varphi / \beta} / \beta)}. \label{eq:lambert_varphi_zi}
\end{align}


\begin{algorithm}[t]
\caption{Proximity Link Generative Models}
\label{algo:gen_process}
\begin{algorithmic}[1]
	\FORALL{edge $e_i$ in $N_s$}
		\STATE Draw an event $z_i \sim Multi(\theta)$
		\STATE Draw a type $t_i \sim Multi(\rho)$
		\STATE Draw a phrase $p_i \sim Multi(\eta_{z_i})$
		\STATE Draw an attribute $x_i \sim Multi(\phi^{t_i}_{z_i})$
	\ENDFOR

	\FORALL{edge $e_j$ in $N_l$}
		\STATE Draw a pair of events $w_{j} \sim Multi(\varphi)$
		\STATE Draw a phrase $y_{j,1} \sim Multi(\eta_{w_{j,1}})$
		\STATE Draw a phrase $y_{j,2} \sim Multi(\eta_{w_{j,2}})$
	\ENDFOR
\end{algorithmic}
\end{algorithm}

\section{Experiments}\label{sec:experiments}

In this section, we evaluate ProxiModel on a variety of news article corpora. We begin by first describing the comparable news corpora we collected for our evaluation, then showing the quality of event descriptors and attributes generated by ProxiModel, when compared to baselines. 
After evaluating the quality of our events, we benchmark the efficiency of our algorithm. We demonstrate the efficiency gains of constructing a compact network for a corpus (without document-level representation) as we increase the number of documents. In addition, we show how using a link minimum support threshold reduces the runtime while maintaining high-quality attributes. Since we have three technical parameters---noise reduction, proximity and sparsity---that affect the quality of event descriptors and attributes as well as method efficiency, we perform parameter studies by varying these parameters to highlight the effects of proximity and sparsity.

Finally, by applying our methodology and extracting key event descriptors and event attributes, we demonstrate how one can construct an event storyline detailing the timeline of events.

\subsection{Datasets}
We evaluate each method on three news corpora, collected from a variety of news agencies through \textsf{NewsBank}~\footnote{\url{www.newsbank.com}}, which cover different distinct topical content.

\begin{itemize}
\parskip -0.2ex
\item \textbf{Sewol Ferry (2014)}:
The sinking of Sewol ferry occured on April 16, 2014, en route from Incheon to Jeju. 
We searched articles with ``Sewol Korea'' keywords, and collected 1,520 articles published from April 15, 2014 to June 30, 2014.

\item \textbf{Japan Tsunami (2011)}:
A massive 8.9-magnitude earthquake shook Japan on March 11, 2011, causing a devastating tsunami to the coast of Japan.
We searched articles with ``Japan Tsunami'' keywords, and collected 21,528 articles published from March 11, 2011 to April 11, 2011.

\item \textbf{Multiple (2014)}:
This dataset has multiple news stories, including Ebola outbreak, the 2014 Winter Olympics, Russian military intervention in Ukraine, missing MH370, Gamboru Ngala attack, Jos bombings, ISIS, Israel-Gaza conflict, and the MH17 tragedy.
We searched articles with multiple keywords for each news story, and collected 100,472 articles published in 2014.
\end{itemize}

Table~\ref{table:datasets} summarizes the collected three datasets. The number of events and the other input parameters can be selected by using cross-validation with perplexity or Bayesian information criterion (BIC)~\cite{schwarz1978estimating}. 
In our study, we set the number of events as follows: 10 for Sewol Ferry, 30 for Japan Tsunami, and 60 for Multiple. As the default values, we set the proximity parameters $\sigma_s$ and $\sigma_l$ to 1 and 10, and the sparsity parameters $\alpha_{L}$, $\alpha_{T}$, $\alpha_{O}$, $\alpha_{P}$, and $\beta$ to 1000, 1000, 10, 10, and 100, respectively.

\begin{table}[t]
\begin{center}
\fontsize{6.5}{7.5}\selectfont
\begin{tabular}{|l||r|r|r|r|r|r|}
\hline
Dataset         & Articles     & Words     & TIME     & GPE       & ORG   & PERSON    \\
\hline
Sewol Ferry     & 1,520         & 5,706     & 67        & 190       & 164       & 235       \\
Japan Tsunami   & 21,528        & 31,793    & 574       & 2,367     & 2,862     & 4,338     \\
Multiple        & 100,472       & 133,540   & 3,565     & 10,907    & 15,417    & 39,093    \\
\hline
\end{tabular}
\end{center}
\vspace{-0.0in}
\caption{Statistics of the datasets: We count words and other entities that appear in at least 5 different news articles.} 
\vspace{-0.05in}
\label{table:datasets}
\end{table}

\subsection{Baselines}

\begin{table*}[t]
\centerline{
    \fontsize{6.5}{7.5}\selectfont
    \begin{tabular} {|l|l|l|l|l|l|}
        \hline
            Method & HISCOVERY & PhraseLDA & ProxiModel-NP & ProxiModel-NS & ProxiModel \\
        \hline
            \multirow{10}{*}{Descriptors}
            & ferry                     & \textbf{third mate}                               & \textbf{save life}            & \textbf{people in need}                       & \textbf{people in need}                       \\
            & ship                      & \textbf{abandon ship}                             & \textbf{third mate}           & \textbf{two crew member}                      & \textbf{two crew member}                      \\
            & people                    & begin list                                        & \textbf{two crew member}      & \textbf{third mate}                           & \textbf{third mate}                           \\
            & \textbf{captain}          & \textbf{senior prosecutor}                        & look whether                  & \textbf{arrest suspicion negligence abandon}  & \textbf{arrest suspicion negligence abandon}  \\
            & passenger                 & 30 minute                                         & distress call                 & look whether                                  & look whether                                  \\
            & \textbf{crew}             & make sharp turn                                   & \textbf{evacuation order}     & \textbf{senior prosecutor}                    & \textbf{evacuation order}                     \\
            & rescue                    & \textbf{arrest warrant}                           & sharp turn                    & \textbf{evacuation order}                     & \textbf{save life}                            \\
            & sewol                     & \textbf{arrest suspicion negligence abandon}      & coast guard                   & news agency                                   & \textbf{arrest warrant}                       \\
            & miss                      & look whether                                      & inside ferry                  & \textbf{save life}                            & \textbf{senior prosecutor}                    \\
            & official                  & \textbf{first mate}                               & high school                   & \textbf{arrest warrant}                       & \textbf{four crew member}                     \\
        \hline
            \multirow{4}{*}{Attributes}
            & SEOUL             & INCHEON         & INCHEON           & MOKPO         & MOKPO         \\
            & 2014-04-20        & 2014-04-19      & 2014-04-19        & 2014-04-19    & 2014-04-19    \\
            & LEE JOON-SEOK     & LEE JOON-SEOK   & LEE JOON-SEOK     & LEE JOON-SEOK & LEE JOON-SEOK \\
            & YANG JUNG-JIN     & YANG JUNG-JIN   & YANG JUNG-JIN     & YANG JUNG-JIN & YANG JUNG-JIN \\
        \hline
    \end{tabular}
}
\caption{Top 10 descriptors from different methods for a key event in Sewol dataset: ``captain arrested on suspicion of negligence'' in 4/19/2014. The descriptors in bold are labeled as key descriptors for the event by at least one participant.}
\label{tbl:descriptors}
\vspace{-0.00in}
\end{table*}

For the comparative study, we have identified two, directly comparable methods and two variations of ProxiModel as baselines for each of our proposed hypotheses.
\begin{itemize}
\parskip -0.2ex
\item \textbf{HISCOVERY}: This work~\cite{li2005probabilistic} assumes each document describes a single event, and the event time is very close to the publication date of the news article. Because of the event time assumption, this model uses publication dates as extra information, which is not available to other baselines.

\item \textbf{PhraseLDA}:  PhraseLDA is proposed in~\cite{elkishky2014topmine}. This model extends Latent Dirichlet allocation to incorporate phrase generation. It utilizes the co-occurrence of phrases or attributes in documents, instead of using proximity. In addition, it has homogeneous outcomes from the generative process, in which all phrases and attributes are generated from a single distribution.

\item \textbf{ProxiModel-NP}:  This is a variation of our model which does not use the proximity information, but instead uses solely the co-occurrence information. It is a special case of ProxiModel, where the proximity parameters ($\sigma$) are set to a very large number. This model serves to see the effectiveness of the proximity information.

\item \textbf{ProxiModel-NS}:  This is our model without the sparse regularization. It is a special case of ProxiModel, where the sparsity parameters are 0. This baseline is designed to show the effects of sparse priors.
\end{itemize}

\subsection{Event Descriptor Evaluation}
In this section we perform a user-study to evaluate the descriptors of the key events across each method.

We select key events from each dataset that can be found across all the methods.

These alignments were performed by expert examination of the descriptors and attributes. One example of a key event used in our evaluation is: ``captain arrested on suspicion of negligence'' in 4/19/2014 which was reported in a news article: ``senior prosecutor Yang Jung-jin said the ferry captain, Lee Joon-seok, 68, faces five charges including negligence of duty and violation of maritime law.''~\footnote{The article can be found in~\url{http://goo.gl/0jW2dO}}

We found 4 events in Sewol Ferry, 10 events in Japan Tsunami, and 16 events in Multiple datasets.
For each key event, we collected the top 10 descriptors from each method, combined and shuffled them to make a method-blind list of descriptors. 
We then asked four participants, who are very familiar with each event and have first read multiple articles for further familiarity, to label each descriptors into the following categories A to C: (A) key descriptors, (B) vague or auxiliary descriptors, and (C) not related. 
The agreement of the labels by the four participants was measured as 0.67 in Fleiss' kappa~\cite{fleiss1971measuring}, indicating substantial agreement.

\begin{figure}
\includegraphics[width=3.4in, angle=0]{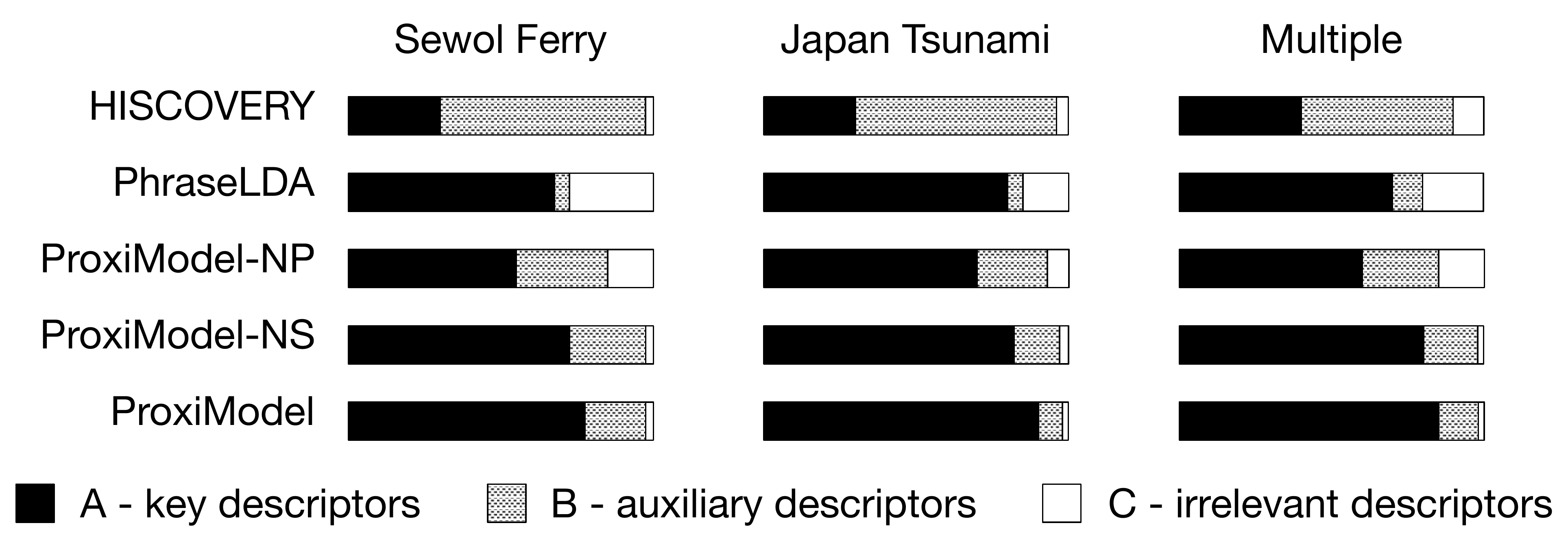}
\vspace{-0.1in}
\caption{The evaluation of descriptors of the aligned key events generated by different methods}
\label{fig:key-descriptors}
\vspace{-0.05in}
\end{figure}

We show an example of aligned key events in Sewol Ferry and the top 10 descriptors and the associated attributes from each methods in Table~\ref{tbl:descriptors}.
Figure~\ref{fig:key-descriptors} shows the distribution of labels for each method.  The phrase-based methods have smaller proportions of B labels than the word-based method, HISCOVERY. In addition, the results show that modeling proximity is important to find key descriptors for events.

\subsection{Event Attribute Evaluation}
\label{sec:auc}

We use a positive and negative set of event attributes for event attribute evaluation. We define a positive set of event attributes as follows:
all attributes in a positive set related to one specific event.

We generated a candidate list of attribute sets and labelled them manually. Table~\ref{tbl:japan_cases} shows some of the annotated event attribute sets.

We compute the probability to generate a given set of attributes from one event as following:
$$
Pr(\tau|\M) = \sum_{e} Pr(e|\M) \prod_{a \in \tau} Pr(a|e),
$$
where $\M$ is a model, $e$ is an event, $\tau$ is a given labeled set of attributes, and $a$ is an attribute in $\tau$.

Based on these probabilities, we rank the labelled sets of attributes to compute the area under the curve (AUC) of the receiver operator curve (ROC)---a curve showing the true positive rate against the false positive rate.  
This is a standard measure used in information retrieval to show the performance of a binary classifier as the discriminatory threshold is varied.  
We can see the performance of our model compared to other baselines in Table~\ref{tbl:auc_eval}. 
While ProxiModel and ProxiModel-NS outperform the other baseline methods, ProxiModel has marginal improvement over ProxiModel-NS. We will address this difference between our sparse model, and non-sparse model in Section~\ref{sec:sparsity}. Also, note that PhraseLDA has lower AUC than ProxiModel, especially in Organization and Person because of using a single distribution for attributes and phrases.

\begin{table}[t]
\fontsize{6.5}{7.5}\selectfont
\centerline{
    \begin{tabular} {|c|c|c|c|}
        \hline
              & Time & Location & Phrase \\
        \hline
            + & 2011-03-11 & HAWAII     & tsunami warning \\
            + & 2011-03-16 & FUKUSHIMA  & nuclear power plant \\
        \hline
            - & 2011-03-11 & CHERNOBYL & cooling system \\
            - & 2011-03-16 & TOKYO     & spend fuel pool \\
        \hline
        \multicolumn{4}{c}{Positive and negative examples of Base}
    \end{tabular}
}
\vspace{0.1in}

\centerline{
    \begin{tabular} {|c|c|c|c|c|}
        \hline
              & Time & Location & Phrase & Org. \\
        \hline
            + & 2011-03-11 & SENDAI     & relief effort   & RED CROSS \\
            + & 2011-03-19 & TOKYO      & radiation level & TEPCO     \\
        \hline
            - & 2011-03-12 & CHERNOBYL  & cooling system  & IAEA      \\
            - & 2011-03-12 & LIBYA      & sweep away      & UN        \\
        \hline
        \multicolumn{5}{c}{Positive and negative examples of Base + Organization}
    \end{tabular}
}
\vspace{0.1in}

\centerline{
    \begin{tabular} {|c|c|c|c|c|}
        \hline
              & Time & Location & Phrase & Person. \\
        \hline
            + & 2011-03-11 & FUKUSHIMA  & stay indoors      & NAOTO KAN \\
            + & 2011-03-17 & FUKUSHIMA  & storage pool      & YUKIO EDANO \\
        \hline
            - & 1979       & UKRAINE    & radioactive material & NAOTO KAN \\
            - & 2011-03-11 & SENDAI     & fuel rod             & BARACK OBAMA \\
        \hline
        \multicolumn{5}{c}{Positive and negative examples of Base + Person}
    \end{tabular}
}
\caption{Examples of human annotated event attributes}
\label{tbl:japan_cases}
\vspace{-0.1in}
\end{table}

\begin{table}[t]
\fontsize{6.5}{7.5}\selectfont
\centerline{
    \begin{tabular} {|c|c|c|c|c|c|}
        \hline
            & HISCOVERY & PhraseLDA & ProxiModel-NP   & ProxiModel-NS   & ProxiModel \\
        \hline
            \multicolumn{6}{|l|}{Sewol Ferry} \\
        \hline
            Base    & 0.5217     & \textbf{0.7971}    & 0.6102    & \textbf{0.8010}  & \textbf{0.8103} \\
            Org.    & 0.5190     & 0.6659             & 0.5111    & \textbf{0.6983}  & \textbf{0.6944} \\
            Person  & 0.5144     & 0.6105             & 0.5308    & \textbf{0.6385}  & \textbf{0.6455} \\
        \hline
            \multicolumn{6}{|l|}{Japan Tsunami} \\
        \hline
            Base    & 0.5149     & 0.6018             & 0.5212    & \textbf{0.6854}  & \textbf{0.6976} \\
            Org.    & 0.4754     & 0.5018             & 0.5594    & \textbf{0.7648}  & \textbf{0.7688} \\
            Person  & 0.6093     & 0.5334             & 0.5291    & \textbf{0.6710}  & \textbf{0.6948} \\
        \hline
            \multicolumn{6}{|l|}{Multiple} \\
        \hline
            Base    & 0.5928     & \textbf{0.7139}    & 0.6272    & \textbf{0.7310}  & \textbf{0.7351} \\
            Org.    & 0.6254     & 0.6740             & 0.6170    & \textbf{0.7564}  & \textbf{0.7431} \\
            Person  & 0.5409     & 0.6688             & 0.6504    & \textbf{0.7605}  & \textbf{0.7660} \\
        \hline
    \end{tabular}
}
\caption{Event retrieval task evaluated using AUC: bold numbers indicate significantly better results than other methods.}
\label{tbl:auc_eval}
\vspace{-0.05in}
\end{table}

\subsection{Parameter Studies}
There are three main parameters in ProxiModel to control the noise reduction: (1) link minimum support, (2) proximity measures, and (3) sparsity of learning parameters. In the following sections, we show how these parameters affect the model's performance.

\subsubsection{Link Minimum Supports}
Because ProxiModel leverages data redundancy, it naturally places higher emphasis on larger link-weights. Taking this into consideration, we apply a minimum support to links in order to reduces the number of trivial links and thus enhance the efficiency of the algorithm. In the Japan tsunami dataset, more than 96\% of links have less than 1.0 weight. By removing small weight links, we have comparable results in quality, but better efficiency.

\begin{figure}[t]
\centering
\hbox {
    \subfigure[AUC for Base]{
    \hspace{-0.14in}
    \includegraphics[width=1.16in, angle=0]{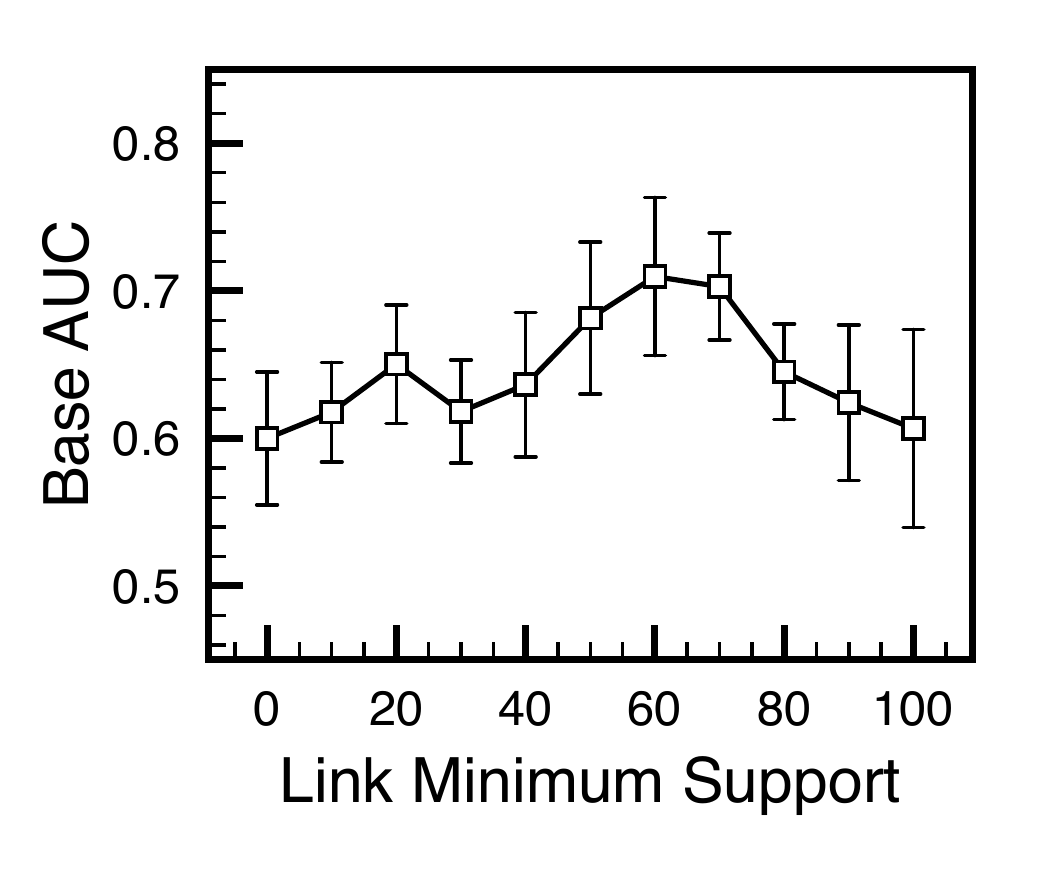}
    \label{fig:linksup_base_auc}
    }
    \subfigure[AUC for Org.]{
    \hspace{-0.14in}
    \includegraphics[width=1.16in, angle=0]{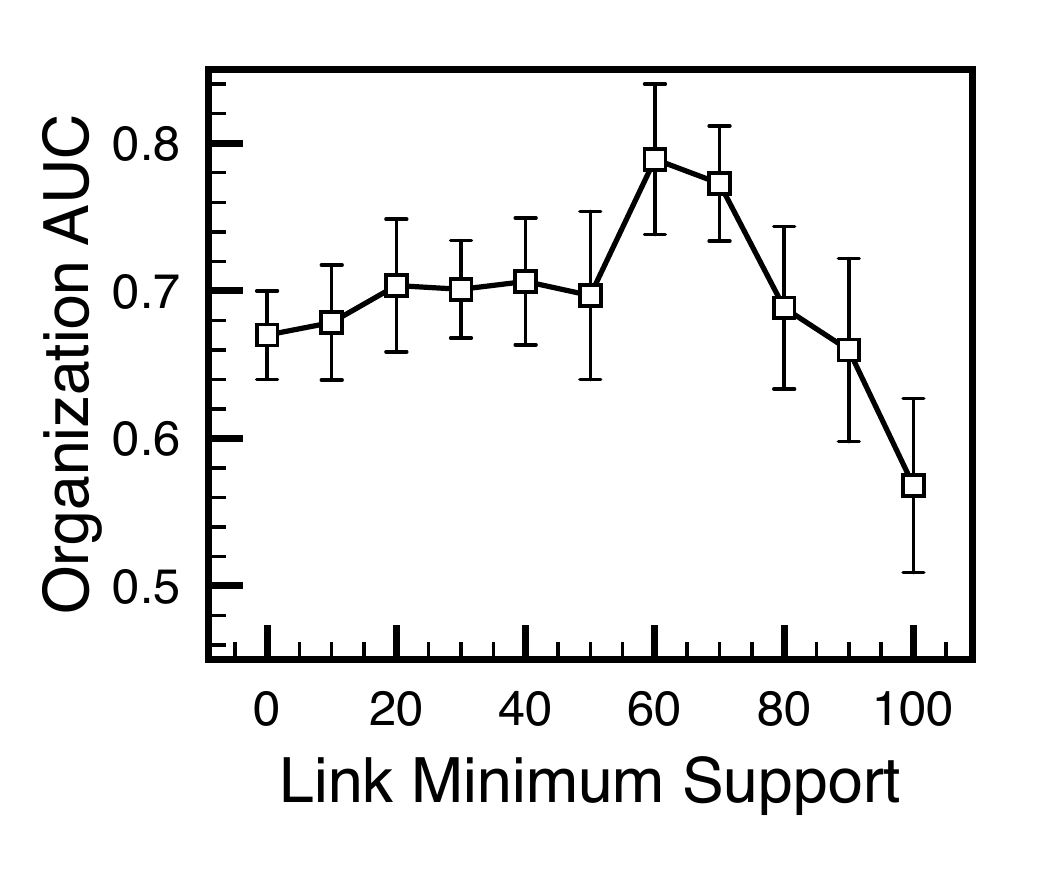}
    \label{fig:linksup_org_auc}
    }
    \subfigure[AUC for Person]{
    \hspace{-0.14in}
    \includegraphics[width=1.16in, angle=0]{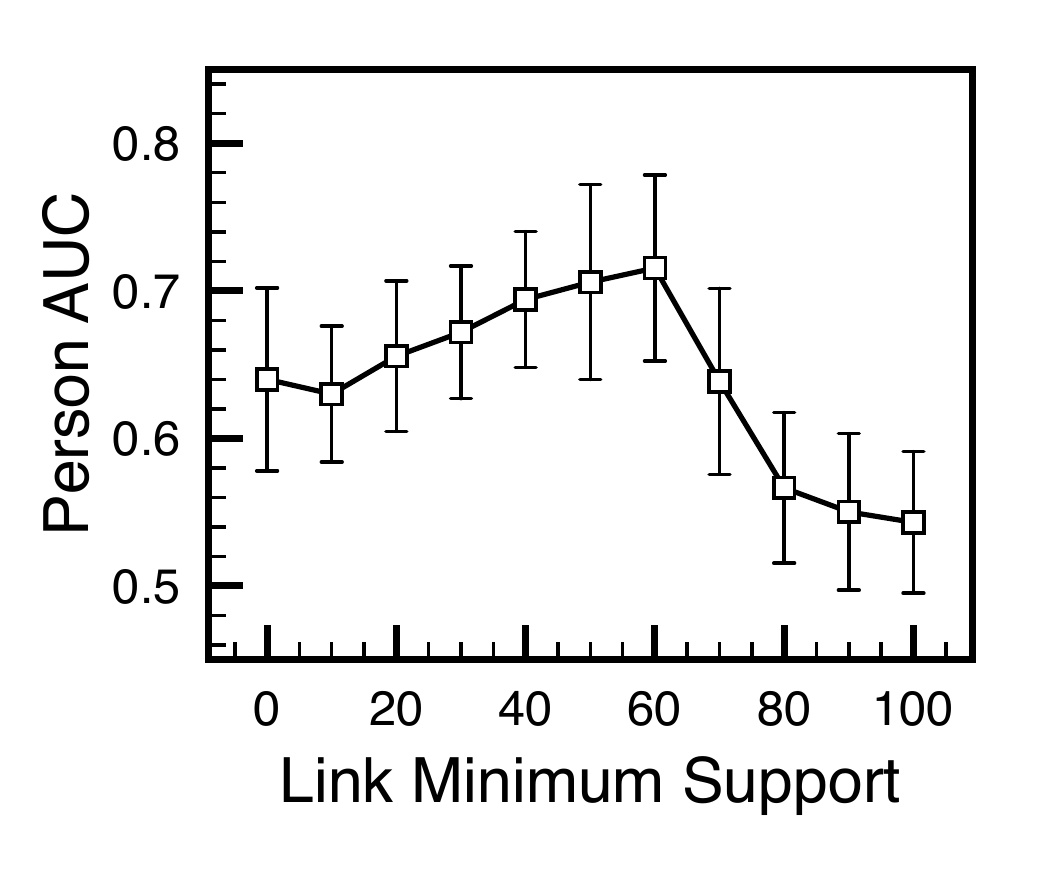}
    \label{fig:linksup_person_auc}
    }
}
\vspace{-0.1in}
\caption{Link Minimum Supports (AUC)}
\label{fig:linksup}
\vspace{-0.05in}
\end{figure}

In Figure~\ref{fig:linksup}, we analyze both our performance as a measure of area under curve of ROC and our runtime performance as we vary the link minimum support parameter. We show the performance of ProxiModel in AUC against different values of link minimum support, $l_{minsup}$. When $l_{minsup}$ is too large, the performance is degraded due to the loss of important information. For all our datasets, we set $l_{minsup}$ to 10.

In Figure~\ref{fig:linksup_running_time}, as we increase the minimum support, proximity networks become sparser, leading to improved efficiency and better runtime.

\subsubsection{Proximity}

\begin{figure}[t]
\centering
\hbox {
    \subfigure[AUC for Base]{
    \hspace{-0.14in}
    \includegraphics[width=1.16in, angle=0]{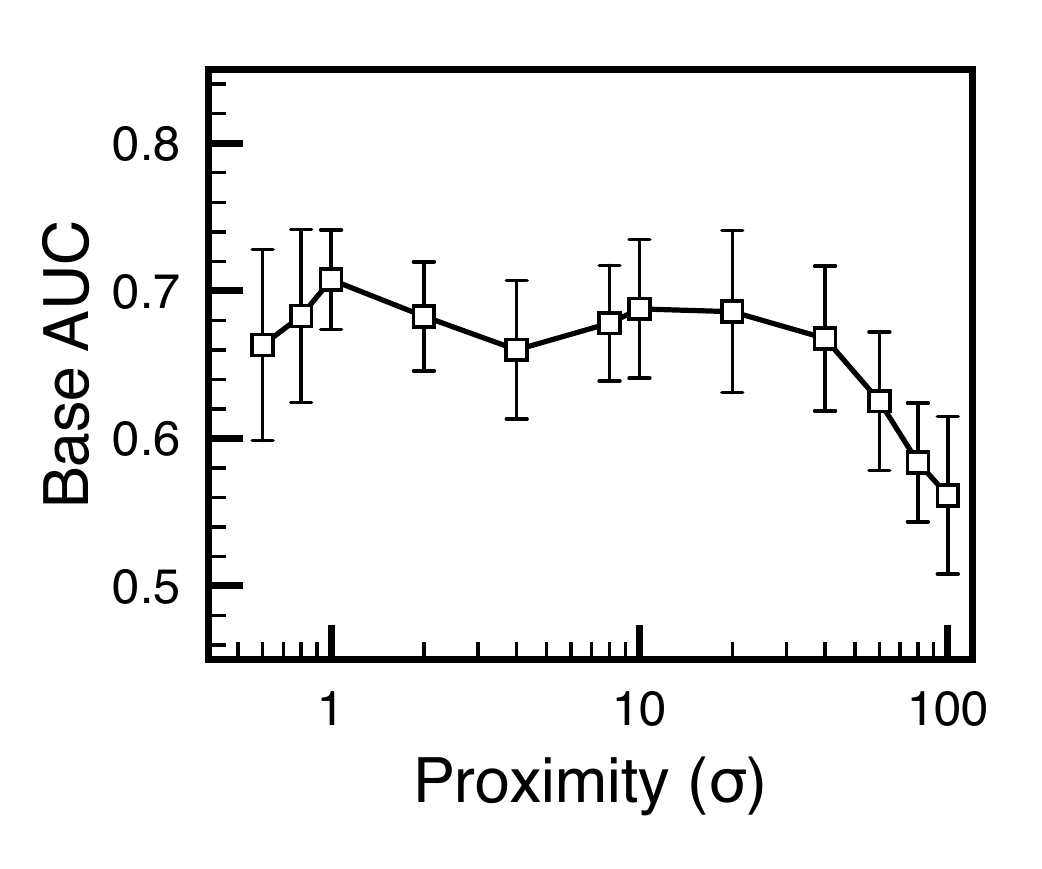}
    \label{fig:proximity_base_auc}
    }
    \subfigure[AUC for Org.]{
    \hspace{-0.14in}
    \includegraphics[width=1.16in, angle=0]{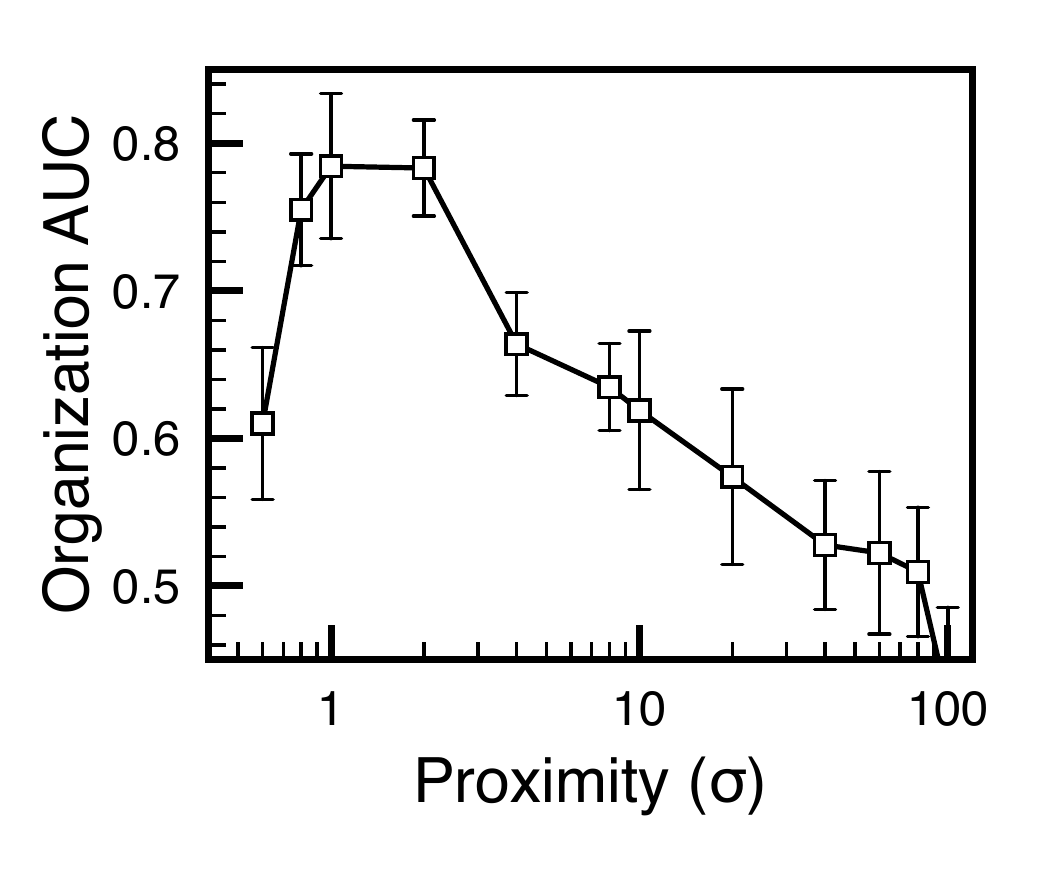}
    \label{fig:proximity_org_auc}
    }
    \subfigure[AUC for Person]{
    \hspace{-0.14in}
    \includegraphics[width=1.16in, angle=0]{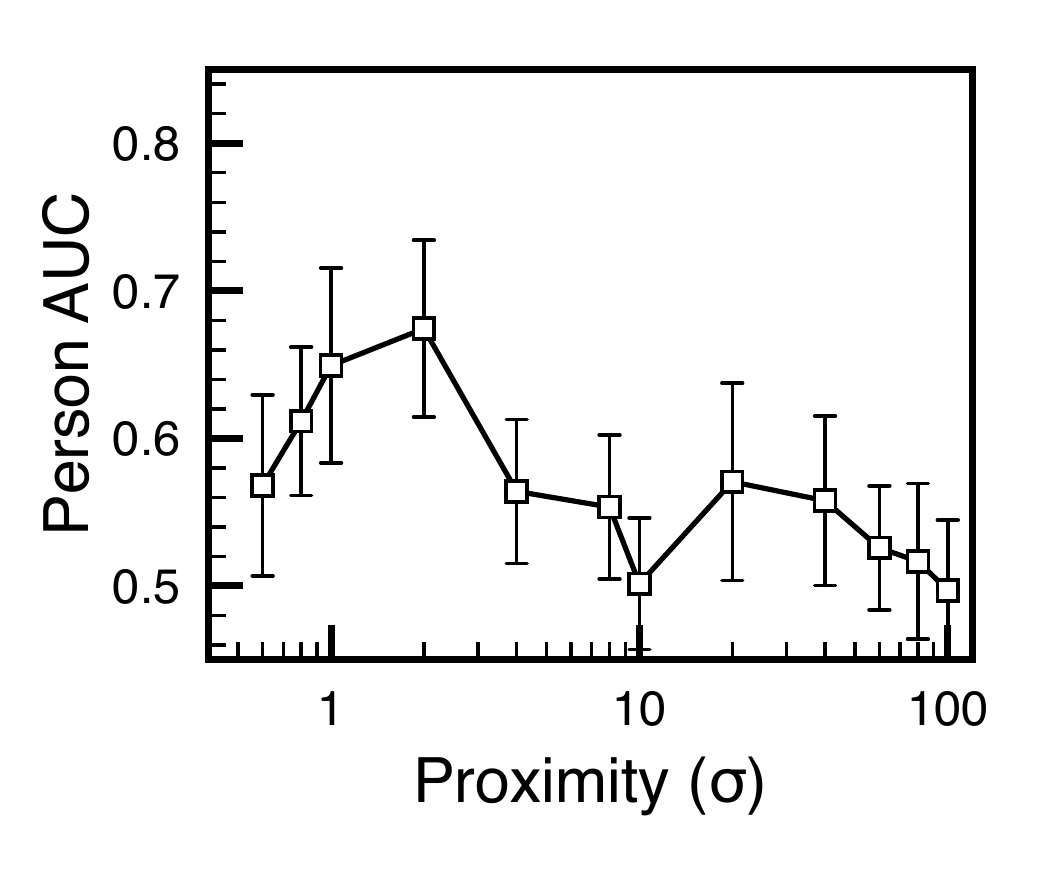}
    \label{fig:proximity_person_auc}
    }
}
\vspace{-0.1in}
\caption{Different $\sigma$ (AUC)}
\label{fig:proximity}
\vspace{-0.1in}
\end{figure}

In Section~\ref{sec:auc}, experiments showed ProxiModel outperforming ProxiModel-NP (non-proximity). In this section we vary $\sigma$ to control our proximity parameter and analyze its effect on retrieval performance. In Figure~\ref{fig:proximity}, we show the performance of our model in AUC with variants with different proximity parameters($\sigma$) for $N_s$. We notice peaks around one in the all figures, but we have significant drops for Organization and Person performance when $\sigma > 2$. As we addressed in Section~\ref{sec:network}, proximity is related to the information propagation within a document.  When $\sigma$ is large, the proximity network captures long range information propagation. For smaller $\sigma$, only near-by information is propagated. Analyzing Figure~\ref{fig:proximity}, we can see indication that for organizations and persons, information is generally propagated in relatively shorter range when compared to location and time information while enjoying long-range propagation. As such, this motivates setting the proximity parameters for each attribute.

\subsubsection{Sparsity}
\label{sec:sparsity}
\begin{figure}[t]
\centering
\hbox {
    \subfigure[Non Sparse Distributions]{
        \hspace{-0.1in}
        \vbox {
            \hbox{ \includegraphics[width=1.65in, angle=0]{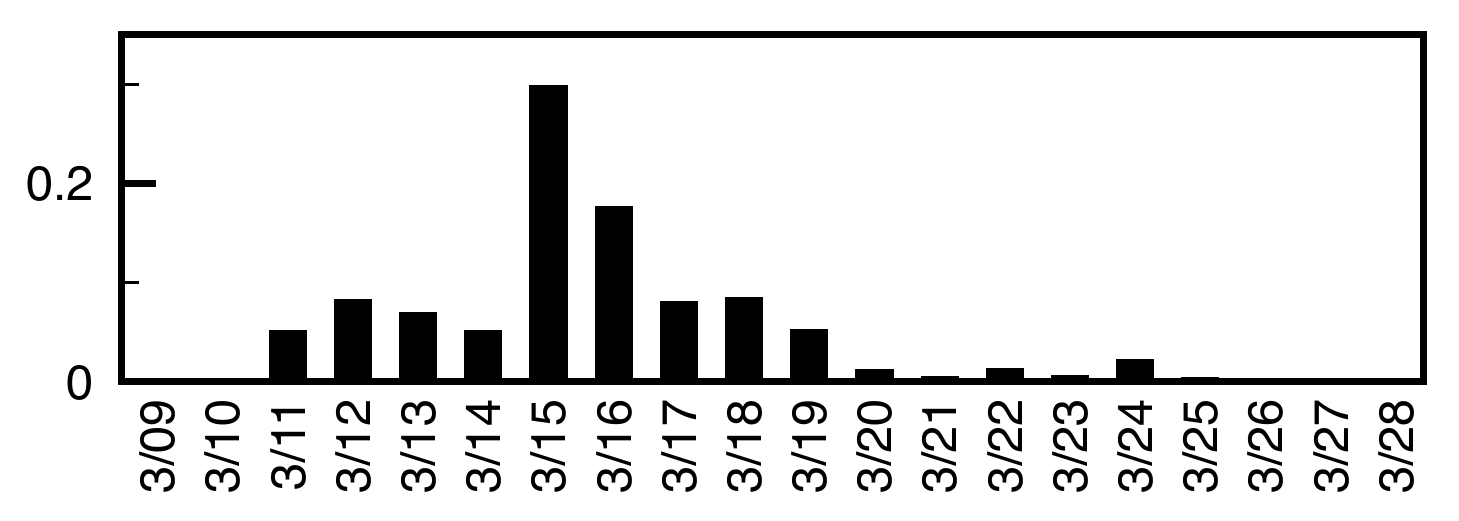}}
            \hbox{ \includegraphics[width=1.65in, angle=0]{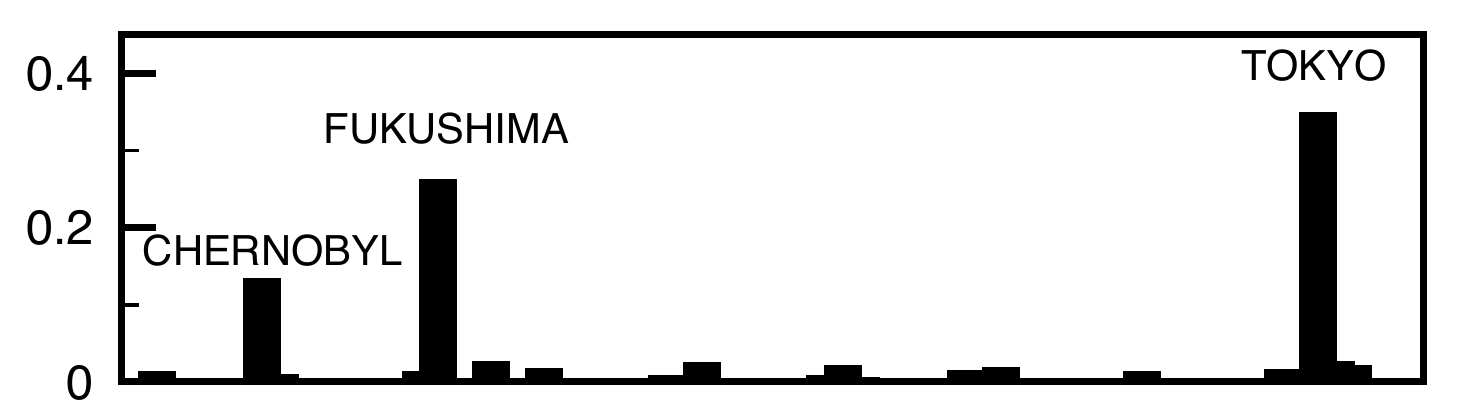}}
        }
        \label{fig:non_sparse_dist}
    }
    \subfigure[Sparse Distributions]{
        \hspace{-0.1in}
        \vbox {
            \hbox{ \includegraphics[width=1.65in, angle=0]{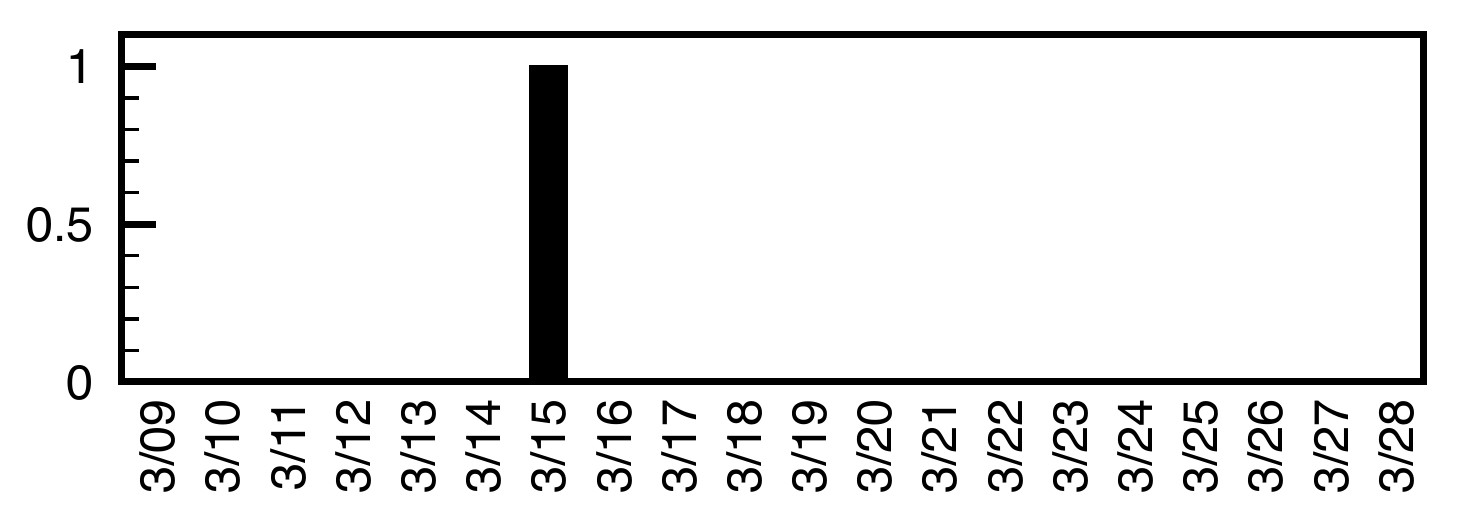}}
            \hbox{ \includegraphics[width=1.65in, angle=0]{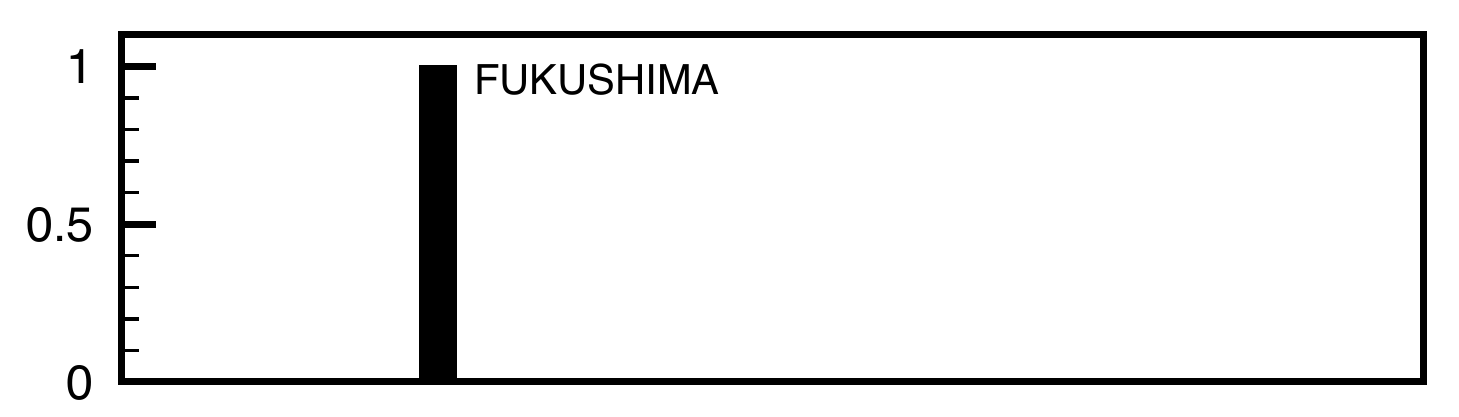}}
        }
        \label{fig:sparse_dist}
    }
}
\vspace{-0.1in}
\caption{Sparsity vs Non-sparsity: Location and Time}
\label{fig:sparsity}
\vspace{-0.05in}
\end{figure}

As mentioned previously, ProxiModel demonstrated marginal improvement over ProxiModel-NS, which was shown to not be statistically significant in Table~\ref{tbl:auc_eval}. While objectively performance is marginal, we observe however that sparsity affects the interpretability of the learned parameters. For example, Figure~\ref{fig:sparsity} shows the learned parameters -- location distribution and time distributions -- for the fire explosion that occurred in the Fukushima nuclear power plant on March 15th. Unlike non-sparse models which display many peaks and thus conflicting information, ProxiModel appears sparse displaying single peaks in the location distribution and time distribution. These are significantly more human-interpretable.

\subsection{Efficiency Analysis}
\begin{figure}[t]
\centering
\hbox {
    \subfigure[Link Min. Support]{
    \hspace{-0.14in}
    \includegraphics[width=1.16in, angle=0]{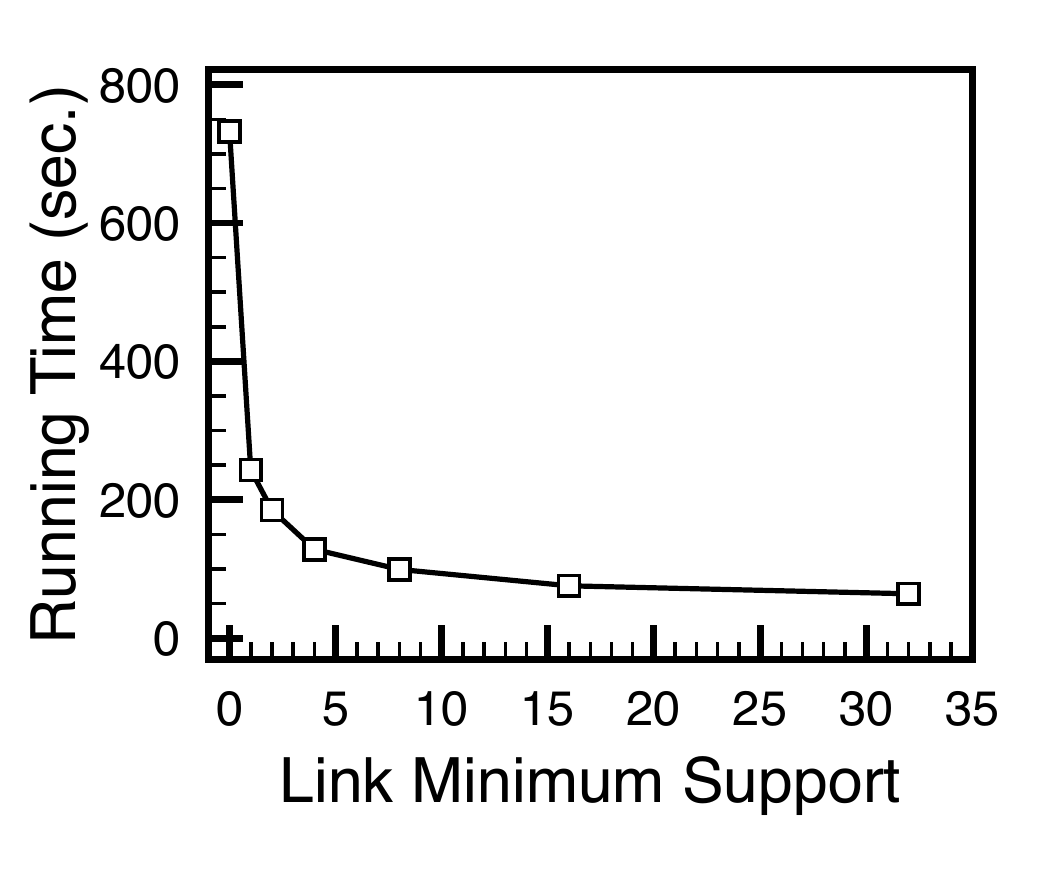}
    \label{fig:linksup_running_time}
    }
    \subfigure[Documents]{
    \hspace{-0.14in}
    \includegraphics[width=1.16in, angle=0]{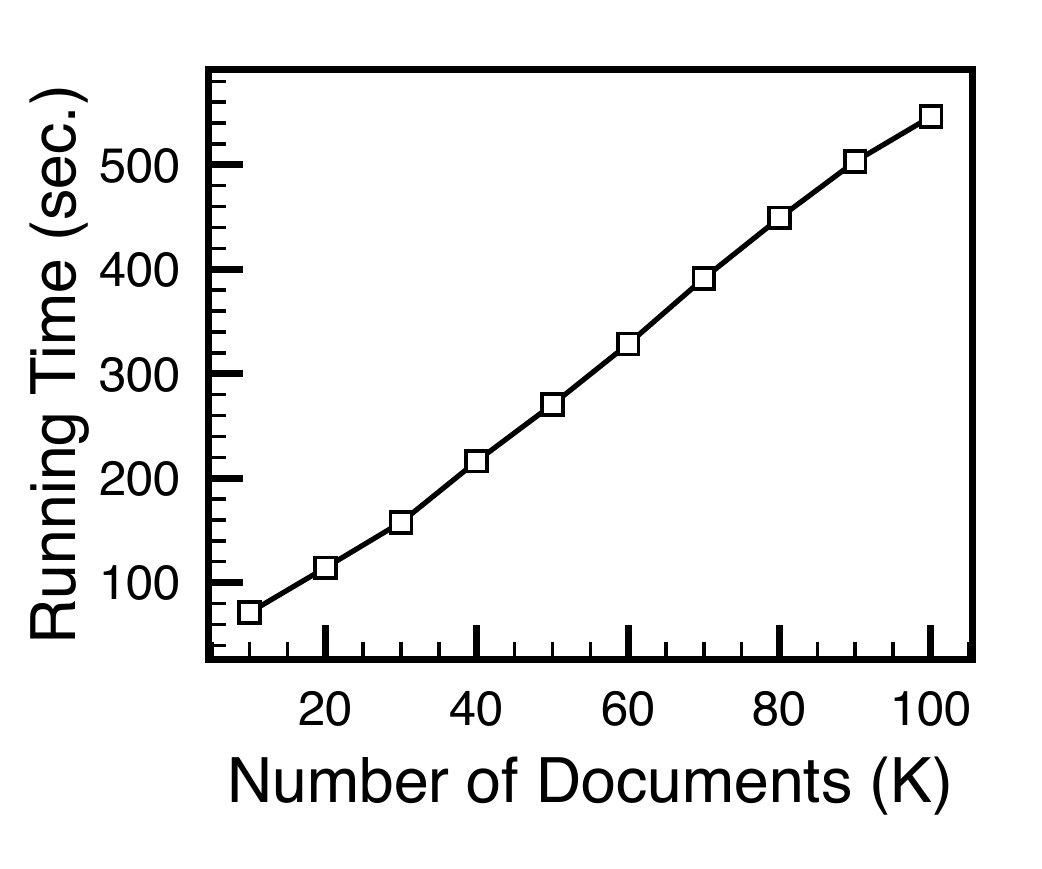}
    \label{fig:docs_running_time}
    }
    \subfigure[Events]{
    \hspace{-0.14in}
    \includegraphics[width=1.16in, angle=0]{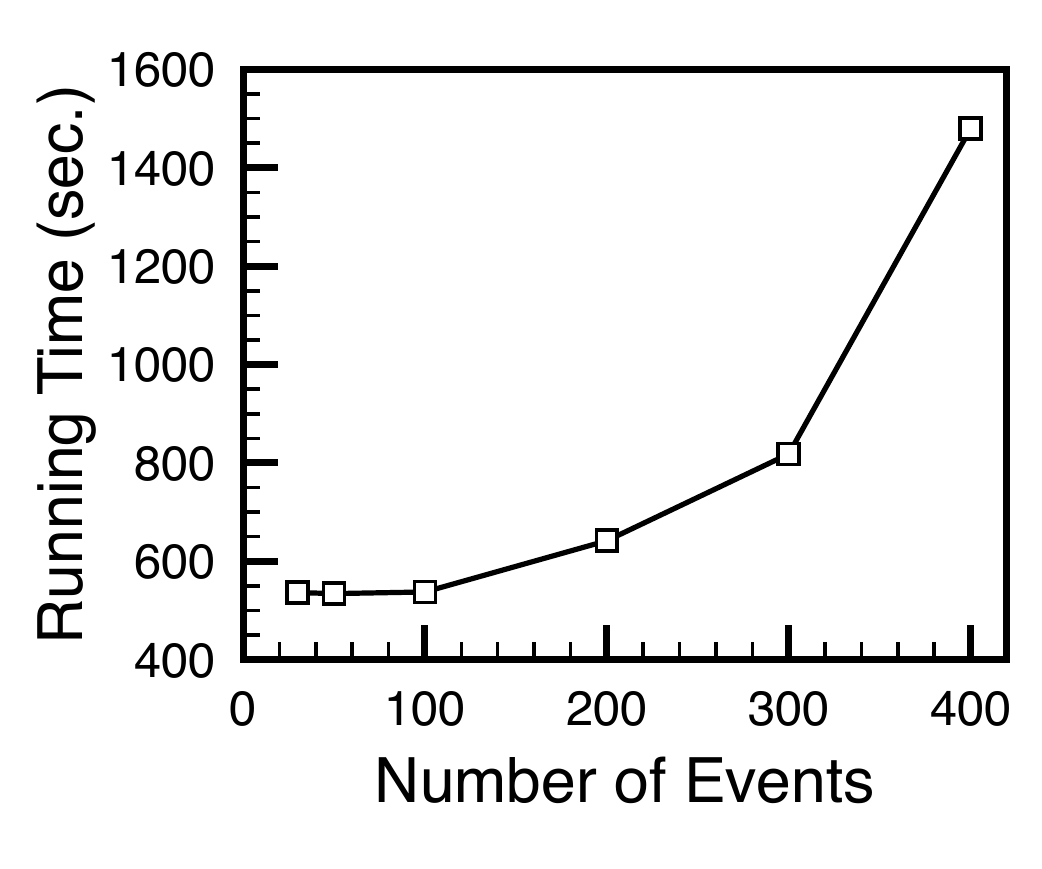}
    \label{fig:events_running_time}
    }
}
\vspace{-0.1in}
\caption{Running Time}
\label{fig:efficiency}
\vspace{-0.1in}
\end{figure}

To understand the run-time efficiency of our methodology, we measure the run-time of ProxiModel using our Multiple dataset, which has approximately 100k documents combined from a variety of sources. We measure runtime as we incrementally increase corpus size. Figure~\ref{fig:docs_running_time} demonstrates empirically run-time is linear in terms of the number of documents. We then vary the number of events parameter and observe run-time performance. From Figure~\ref{fig:events_running_time} we can see that runtime is quadratic in relation to number of events. As this parameter is usually small (a small number of events), this is less significant  than linearity with respect to corpus size.

\subsection{Visualization}

We use ProxiModel to learn Japan Tsunami events and their connections, and visualize them in Figure~\ref{fig:big-picture}. 
An event is represented by a circle with a radius proportional to its probability in event distribution $\theta$. 
Each event is described by a list of top 6 event descriptors from $\eta_z$ in conjunction with top event attributes (e.g., time, location, organization, persons) from $\phi$. 
For some events, there could be no relevant event attributes for a certain type. 
When $\sum_i \hat{e}^{s,t}_{i,j,z} < 1$ for a top event attribute $j$ of an event $z$, the top event attribute is ignored and shown as --. 
This combination of human-interpretable, multi-word phrases with event attributes help an individual to better understand each event.

In addition, links between events are drawn based on $\varphi$. 
Since we impose sparsity on $\varphi$, there are only a few non-trivial links between events. 
Each line width is proportional to its probability in event link distribution $\varphi$. 
The links between events help chain related events together naturally forming an easy-to-interpret branching timeline story. 
By systematically traversing this event graph, one can naturally construct a storyline of the significant events surrounding the Japan nuclear disaster.

\section{Related Work}\label{sec:related}

With the explosion of digitalized news data, identifying and organizing news articles into events has emerged as a key method of improving access and reuse of these large news collections. 
Many attempts have been made to extract events from text corpora. These approaches can be categorized into NLP-based contextual analysis approaches and data mining approaches.

In the NLP literature, many approaches employ rich features to model event extraction as a parsing problem. McClosky \textit{et al.} perform event extraction by creating a tree of event-argument relations and using this as a representation for reranking of the dependency parser~\cite{mcclosky2011event}. NLP event extraction techniques have even been applied to extracting biomedical events from text literature such as binding, regulation, and gene-protein interactions; these techniques rely on a rich feature-set for classification~\cite{miwa2010event}. Other methods employ tagging and matching specified event patterns to perform large-scale event extraction; redundancy is reduced by automatically generating rulesets for event merging~\cite{aone2000rees}. While these NLP-based methods often obtain high-quality results, their dependency on parsing, user-defined patterns, and annotated data reduces effectiveness across multiple sources. While these methods may show acceptable performance in a closed-domain such as when the types of events are known before-hand, they suffer in an open-domain scenario.

In the data mining literature, a variety of methods have been introduced for extracting underlying events from news corpora. Using a probabilistic model that incorporates both content, time, and location information, Li \textit{et al.} develop a unified generative model where, for each article, a single latent event generates observable event descriptors such as location, people, keywords and timestamps~\cite{li2005probabilistic}. This HISCOVERY framework first applies NLP entity recognition tools to extract persons, locations, and dates/times, then uses this data in its generative model. However, it makes the strong assumption that each news article references a single event, a requirement we relax in our probabilistic model. Other works have purely focused on extending topic models for identifying phrases, topical hierarchies, and entities in news corpora~\cite{han2014bringing,ren2016automatic2}.

Other approaches in the data mining literature apply clustering and document relevancy measures to organize documents into coherent events. 
These method often employ heuristic clustering approaches based on intra-cluster similarity to agglomeratively form event clusters. 
Naughton \textit{et al.} annotate sentences with event labels then aggregate these sentences into a structured form and create coherent event summaries~\cite{naughton2006clustering}.
They also apply machine learning to extract event-containing sentences and propose two metrics for event sentence clustering to identify, integrate, and summarize news events from multiple sources~\cite{naughton2006event}.
Further clustering approaches agglomeratively merge and prune event clusters to identify discriminative events~\cite{sayyadi2009event}. 
Lam \textit{et al.} cluster documents into events and detect new events by first extracting discriminative ``concept terms", named entities, and other identifying information and using these features, cluster documents into existing and new events~\cite{lam2001using}. 
These clustering approaches are document-level event analysis, defining an event as a collection of topically related article. 
These works are not suitable for fine-grained event analysis.  

\vspace*{-1.2ex}
\section{Conclusions}
\label{sec:conclusion}

In this study, we address the problem of, given large, noisy comparable news corpora, extracting events, their identifying attributes along with  interpretable descriptors. We design a novel event mining framework to integrate phrases, named entities, and time expressions to construct then cluster proximity networks to identify these hidden events. A key aspect of our approach involves utilizing proximity of information consistently found in corpus in order to model and propagate event information. By evaluating our approach on three news corpora with different topical contents, we validate our ability to generate concise, accurate, and interpretable key descriptors and attributes consistently beating the comparable baselines. 

\section{Acknowledgements}
Research was sponsored in part by U.S. Army Research Lab. under Cooperative Agreement No. W911NF-09-2-0053 (NSCTA), DARPA under Agreements No. W911NF-17-C-0099 and FA8750-19-2-1004, National Science Foundation IIS 16-18481, IIS 17-04532, and IIS-17-41317, DTRA HDTRA11810026, and grant 1U54GM114838 awarded by NIGMS through funds provided by the trans-NIH Big Data to Knowledge (BD2K) initiative (www.bd2k.nih.gov). Any opinions, findings, and conclusions or recommendations expressed in this document are those of the author(s) and should not be interpreted as the views of any U.S. Government. The U.S. Government is authorized to reproduce and distribute reprints for Government purposes notwithstanding any copyright notation hereon.

\balance
\bibliographystyle{plain}
\bibliography{big_data_19}

\end{document}